\documentclass[runningheads]{llncs}
\usepackage{graphicx}
\usepackage{comment}
\usepackage{amsmath} 
\usepackage{amssymb}
\usepackage{color}

\usepackage{amsfonts}       
\usepackage{nicefrac}       
\usepackage{microtype}      
\usepackage{xcolor}
\usepackage{multirow}
\usepackage{authblk}
\usepackage{tabularx}

\usepackage{makecell} 
\usepackage{bm} 
\usepackage{subfigure}

\begin{document}
\pagestyle{headings}
\mainmatter
\def\ECCVSubNumber{5633}  

\title{Multi-Domain Image Completion for \\ Random Missing Input Data} 


\titlerunning{Multi-Domain Image Completion}
%
\author{Liyue Shen\inst{1}\footnote{This work was done when L. Shen was an intern at NVIDIA.} \and Wentao Zhu\inst{2} \and Xiaosong Wang\inst{2} \and Lei Xing\inst{1} \and John M. Pauly\inst{1} \and Baris Turkbey\inst{3} \and Stephanie Anne Harmon\inst{3} \and Thomas Hogue Sanford\inst{3} \and Sherif Mehralivand\inst{3} \and Peter Choyke\inst{3} \and Bradford Wood\inst{3} \and Daguang Xu\inst{2}}
\authorrunning{L. Shen et al.}
%
\institute{Stanford University \email{\{liyues,lei,pauly\}@stanford.edu} \and
NVIDIA \email{\{wentaoz,xiaosongw,daguangx\}@nvidia.com} \and
National Institutes of Health}
\maketitle

\begin{abstract}
Multi-domain data are widely leveraged in vision applications taking advantage of complementary information from different modalities, e.g., brain tumor segmentation from multi-parametric magnetic resonance imaging (MRI). However, due to possible data corruption and different imaging protocols, the availability of images for each domain could vary amongst multiple data sources in practice, which makes it challenging to build a universal model with a varied set of input data. To tackle this problem, we propose a general approach to complete the random missing domain(s) data in real applications. Specifically, we develop a novel multi-domain image completion method that utilizes a generative adversarial network (GAN) with a representational disentanglement scheme to extract shared ``skeleton'' encoding and separate ``flesh'' encoding across multiple domains. We further illustrate that the learned representation in multi-domain image completion could be leveraged for high-level tasks, e.g., segmentation, by introducing a unified framework consisting of image completion and segmentation with a shared content encoder. The experiments demonstrate consistent performance improvement on three datasets for brain tumor segmentation, prostate segmentation, and facial expression image completion respectively.

\end{abstract}

\section{Introduction}
Multi-domain images are often required as inputs in various vision tasks because of the nature that different domains could provide complementary knowledge. For example, four medical imaging modalities, MRI with T1, T1-weighted, T2-weighted, FLAIR (FLuid-Attenuated Inversion Recovery), are acquired as a standard protocol to accurately segment the tumor regions for each patient in the brain tumor segmentation task~\cite{brats2014tmi}. Different modalities provide distinct features to locate tumor boundaries from differential diagnosis perspectives. Additionally, when it comes to the natural image tasks, there are similar scenarios such as person re-identification across different cameras or times~\cite{zheng2015market1501,zheng2019jointreid}. Here, the medical images in different modalities or natural images with the person under varied appearances can be considered as different image domains, depicting the same underlying subject or scene from various aspects.

\begin{figure}[t]
\centering
\includegraphics[width=0.7\linewidth]{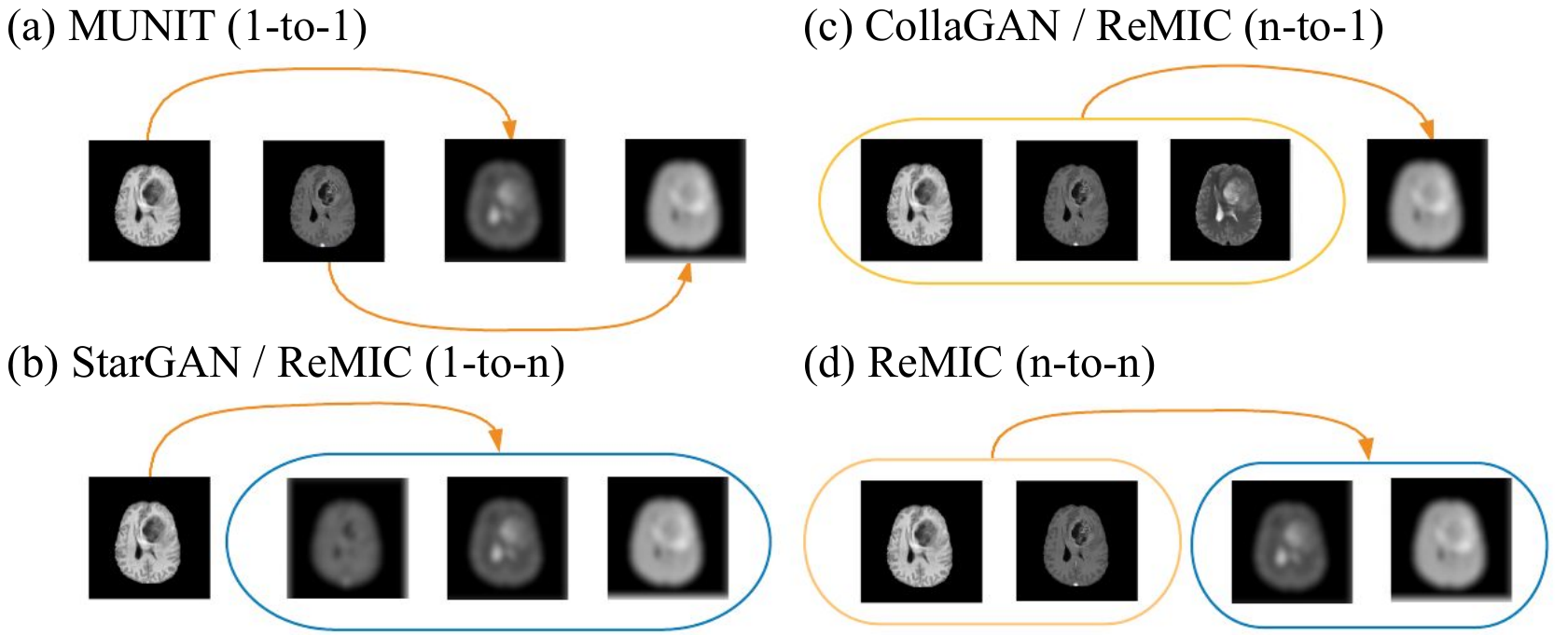}
\caption{Image translation using (a) MUNIT (1-to-1), (b) StarGAN / Ours (ReMIC) (1-to-$n$), (c) CollaGAN / ReMIC ($n$-to-1), and (d) ReMIC ($n$-to-$n$). In multi-domain image completion, Ours (ReMIC) completes the missing-domain images given randomly distributed numbers ($k$-to-$n$, $1 \leq k \leq n$) of visible domains in the input. Note the missing-domain images are denoted as blurred images.}
\label{fig:motivation}
\vspace{-0.1in}
\end{figure}

However, some image domains might be missing in practice. Especially when it comes to a  large-scale multi-institute study, it is generally difficult or even infeasible to guarantee the availability of data in all domains for every data entry. For example, some patients might lack certain imaging scans due to different imaging protocols, data loss or image corruption. For these rare and valuable collected data, it is costly to just throw away the incomplete samples during training, and also infeasible to test with missing-domain inputs. Thus, in order to take the most advantage of such missing data, it becomes crucial to design an effective data completion algorithm to cope with this challenge. An intuitive approach is to impute the missing domain of one sample with the nearest neighbor from other samples whose corresponding domain image exists. But this might lack of semantic consistency among different domains of the input sample as shown in Fig. 2 since it only focuses on the pixel-level similarity compared with existing images. Another possible solution is to generate images and complete missing domains via image translation from existing domains using generative models, such as GAN models, as illustrated in Fig.~\ref{fig:motivation}. 

In this work, we propose a general \(n\)-to-\(n\) image completion framework based on a Representational disentanglement scheme for Multi-domain Image Completion (ReMIC).
Specifically, our contribution is fourfold: (1) We propose a novel GAN framework for a general and flexible \(n\)-to-\(n\) image generation with representational disentanglement, i.e., learning semantically shared representations cross domains (content code) and domain-specific features (style code) for each input domain; (2) We demonstrate the learned content code could be utilized for the high-level task, i.e., developing a unified framework for jointly learning the image completion and segmentation based on shared content encoder; (3) We demonstrate the proposed \(n\)-to-\(n\) image generation model can effectively completes the missing domains given randomly distributed numbers ($k$-to-$n$, $1 \leq k \leq n$) of visible domains in the input; (4) Experiments on three datasets illustrate that the proposed method consistently achieves better performance than previous approaches in both multi-domain image completion and missing-domain segmentation.

\begin{figure}[t]
\centering
    {\includegraphics[width=0.7\linewidth]{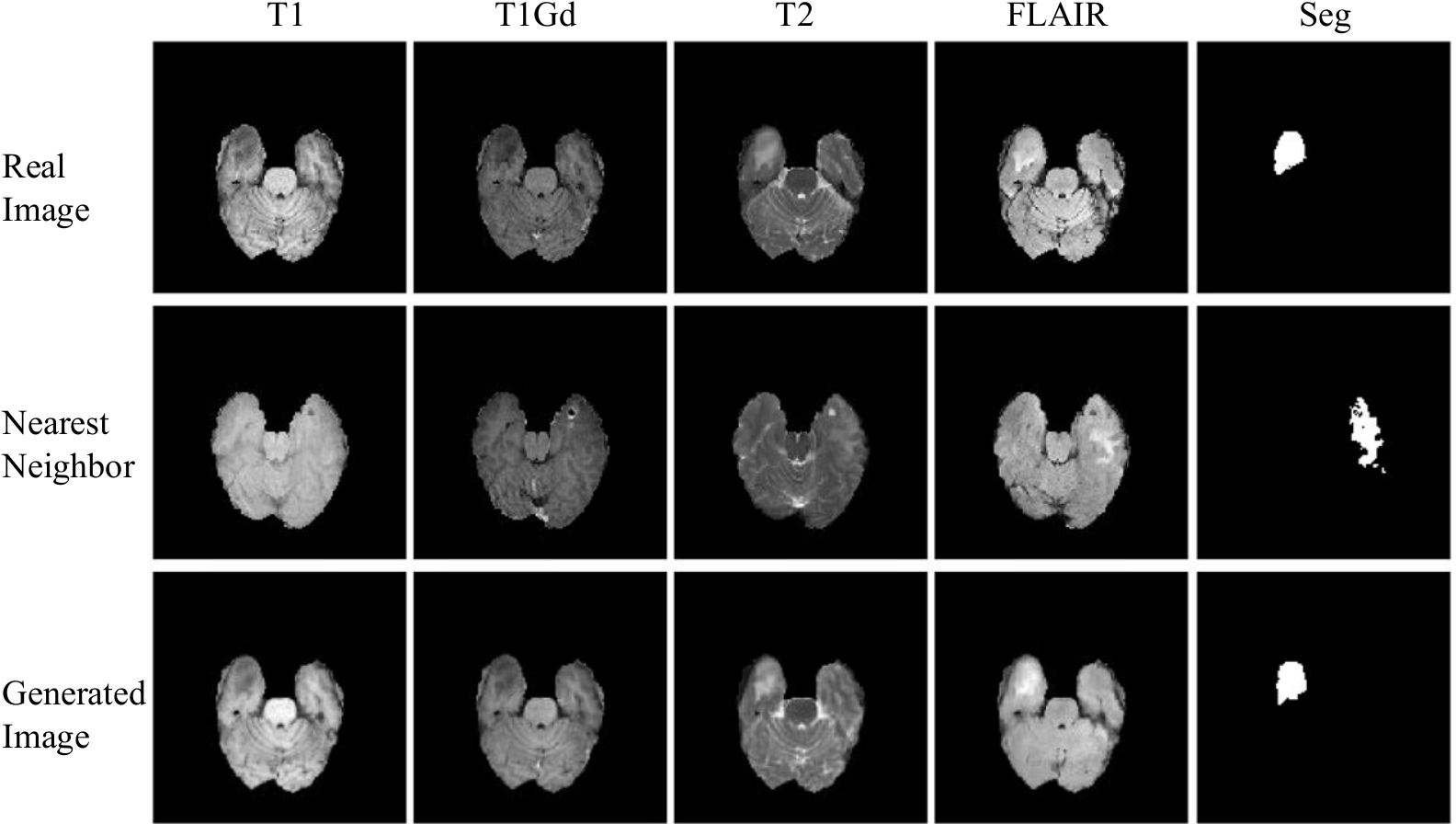}}
    {\caption{BraTS images in four modalities with nearest neighbors and generated images from the proposed method (ReMIC). From the segmentation prediction of brain tumor, the generated images preserve better semantic consistency with ground truth in addition to the pixel-level similarity in images.}}
\label{fig:nn_seg}
\vspace{-0.1in}
\end{figure}

\section{Related Work}

\noindent \textbf{Image-to-Image Translation} The recent success of GANs~\cite{goodfellow2014generative,mirza2014conditionalgan,isola2017pix2pix,zhu2017cyclegan,zhu2017bicyclegan,kim2017discogan,liu2017unit,choi2018stargan,yoon2018radialgan,lee2019collagan,zhang2018mri_ct_translating,dar2019MMRI_gan,Sharma2019MissingMP} in image-to-image translation provides a promising solution to deal with the challenge of missing image domains. CycleGAN~\cite{zhu2017cyclegan} shows impressive performance in image-to-image translation via cycle-consistency between real and generated images. 
However, it mainly focuses on 1-to-1 mapping between two domains and assumes corresponding images in two domains strictly share the same representation in latent space. This is limited in multi-domain applications since $\frac{n(n-1)}{2}$ CycleGAN models are required if there are $n$ domains. Following this, StarGAN~\cite{choi2018stargan} proposes to use a mask vector in inputs to specify the desired target domain in multi-domain image generation. Meantime, RadialGAN~\cite{yoon2018radialgan} also deals with the multi-domain generation problem by assuming all the domains share the same latent space. Although these works make it possible to generate images in different target domains through 1-to-\(n\) mapping with multiple inference passes, the representation learning and image generation are always conditioned on the single input image as the only source domain. In order to take advantage of multiple available domains, CollaGAN~\cite{lee2019collagan} proposes a collaborative model to incorporate multiple domains for generating one missing domain. Similar to StarGAN, CollaGAN relies on the cycle-consistency to preserve the contents in the generated images, which is an indirect and implicit constraint for target domain images. Additionally, since the target domain is specified by an one-hot mask vector in input, CollaGAN is essentially doing \(n\)-to-1 translation with a single output in an one-time inference. As illustrated in Fig.~\ref{fig:motivation}, our proposed model is a more general \(n\)-to-\(n\) image generation framework that can overcome aforementioned limitations.

\noindent \textbf{Learning Disentangled Representations}
Recently, learning disentangled representations is proposed to capture the full distribution of possible outputs by introducing a random style code~\cite{chen2016infogan,higgins2017betavae,huang2018munit,lee2018drit,lee2019drit++,lin2019dosgan}, or to transfer information across domains for adaptation~\cite{liu2018detach,liu2018unified}. InfoGAN~\cite{chen2016infogan} and $\beta$-VAE~\cite{higgins2017betavae} learn the disentangled representation in an unsupervised manner. In image translation, DRIT~\cite{lee2018drit} disentangles content and attribute features by exchanging the features encoded from two domains respectively. The image consistency during translation is constrained by the code and image reconstruction. With a similar code exchange scheme, MUNIT~\cite{huang2018munit} assumes a prior distribution on style code, which allows directly sampling style codes from the prior distribution to generate target domain images. However, both DRIT and MUNIT only deal with image translation between two domains, which requires to independently train $\frac{n(n-1)}{2}$ separate translation models among $n$ domains. While the recent work~\cite{liu2018unified} also tackles multi-domain image translation, it focuses more on learning cross-domain latent code for domain adaptation with less discussion about the domain-specific style code. Moreover, our proposed method handles a more challenging problem with random missing domains motivated by practical medical applications. Aiming at higher completion accuracy for the segmentation task with missing domains, we further add reconstruction and segmentation constraints in our framework.

\noindent \textbf{Medical Image Synthesis}
Synthesizing medical images has attracted increasing interests in recent researches~\cite{zhang2018mri_ct_translating,dar2019MMRI_gan,Sharma2019MissingMP,huo2018adversarial,iglesias2013synthesizing,shrivastava2017learning,costa2017towards,kamnitsas2017unsupervised,nie2017medical,zhu2018adversarial}. The synthesized images are generated across multi-contrast MRI modalities or between MRI and computed tomography (CT).~\cite{van2015cross,havaei2016hemis,chartsias2017multimodal} also discuss how to extract representations from multi-modalities especially for segmentation with missing imaging modalities. However, these studies mostly focus on how to fuse the features from multiple modalities but not from the perspective of representation disentanglement. Our model disentangles the shared content and separate style representations for a more general \(n\)-to-\(n\) multi-domain image completion task, and we further validate that the generation benefits the segmentation task.



\section{Method}

\begin{figure}[t]
\centering
{\includegraphics[width=0.8\linewidth]{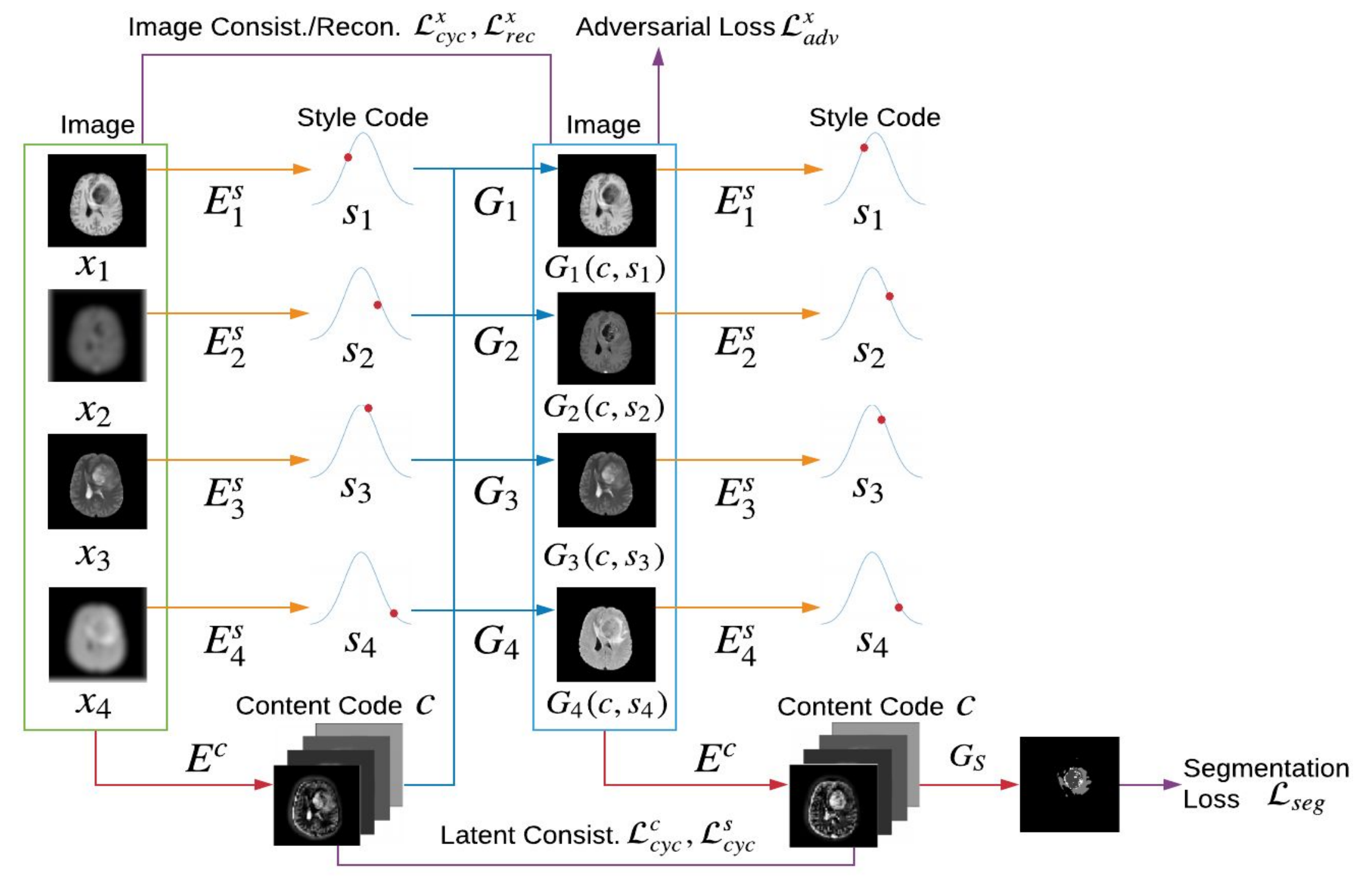}}
  {\caption{Overview of the proposed \(n\)-to-\(n\) multi-domain completion and segmentation framework. $N=4$ and two domains ($x_2, x_4$) are missing in this example. Our model contains a unified content encoder $E^c$ (red lines), domain-specific style encoders $E^s_i$ (orange lines) and generators $G_i$ (blue lines), $1 \leq i \leq N$. A variety of losses are adopted (burgundy lines), i.e., image consistency loss for visible domains $\mathcal{L}_{cyc}^x$, latent consistency loss $\mathcal{L}_{cyc}^c$ and $\mathcal{L}_{cyc}^s$, adversarial loss $\mathcal{L}_{adv}^x$ and reconstruction loss $\mathcal{L}_{rec}^x$ for the generated images. Furthermore, representational learning framework combines a segmentation generator $G_S$ following the content code for a unified image generation and segmentation.}
  }
\label{fig:model}
\vspace{-0.1in}
\end{figure} 

Images from different domains for the same sample present their exclusive features of the subject. Nonetheless, they also inherit some global content structures. For instance, in the parametric MRI for brain tumors, T2 and FLAIR MRI highlight the differences in tissues' water relaxational properties, which will distinguish the tumor tissue from normal ones. Contrasted T1 MRI can examine the pathological intratumoral take-up of contrast agents so that the boundary between tumor core and the rest will be highlighted. However, the underlying anatomical structure of the brain is shared by all these modalities. With the availability of multiple domain data, it is meaningful to decompose the images into the shared content structure (skeleton) and their unique characteristics (flesh) through learning. Therefore, we will be able to reconstruct the missing image during the testing by using the shared skeleton (extracted from the available data domains) and a sampled flesh from the learned model. Without assuming a fixed set of missing domains during the training, the learned framework could flexibly handle one or more missing domains in a random set. In addition, we further enforce the accuracy of the extracted content structure by connecting it to the segmentation task. In such manner, the disentangled representations of multiple domain images (both the skeleton and flesh) can help both the image completion and segmentation.

Suppose there are $N$ domains: $\{ \mathbf{\chi}_1, \mathbf{\chi}_2, \cdots, \mathbf{\chi}_N \}$. Let $x_1 \in \mathbf{\chi}_1$, $x_2 \in \mathbf{\chi}_2$, $\cdots$, $x_N \in \mathbf{\chi}_N$ be the images from $N$ different domains respectively, which are grouped data describing the same subject $\mathbf{x}=\{x_1, \cdots, x_N\}$ as one sample. Assume the dataset contains $M$ independent data samples in total. For each sample, we assume one or many of the $N$ domain images might be randomly missing, i.e. the number and category of missing domains are both random. The goal of our first task is to complete all the missing domains for a random input sample.

To accomplish the completion of all missing domains from a random set of available domains, we assume the $N$ domains share the latent representation of underlying structure. We name the shared latent representation as \textit{content code} and meanwhile each domain also exclusively contains the domain-specific latent representation, i.e., \textit{style code}, that is related to various characteristics or attributes in different domains. 
The missing domains can be reconstructed from these two aspects of information through the learning of deep neural networks. Similar to the setting in MUNIT~\cite{huang2018munit}, we assume a prior distribution for style latent code as $\mathcal{N}(\bm{0},\bm{I})$ to capture the full distribution of possible styles in each domain. However, MUNIT trains separate content encoder for each domain and enforce the disentanglement via coupled cross-domain translation during training while our method employs a single content encoder to extract the anatomic representation shared across all the domains.

\subsection{Unified Image Completion and Segmentation}
As shown in Fig.~\ref{fig:model}, our model contains a unified content encoder $E^c$ and domain-specific style encoders $E^s_i$ ($1 \leq i \leq N$), where $N$ is the total number of domains. Content encoder $E^c$ extracts the shared content code $c$ from all existing domains: $E^c(x_1, x_2, \cdots, x_N) = c$. For the missing domains, we use zero padding in corresponding input channels. For each domain, a style encoder $E^s_i$ learns the domain-specific style code $s_i$ from the corresponding domain image $x_i$ ($1 \leq i \leq N$) respectively: $E^s_i(x_i) = s_i$.

During the training, our model captures the shared content code $c$ and separate style codes $s_i$ ($1 \leq i \leq N$) through the disentanglement process (denoted as red and orange arrows respectively in Fig.~\ref{fig:model}) with a random set of input images (in green box). In Fig.~\ref{fig:content_brats}, we visualize the extracted content codes (randomly selected 8 out of 256 channels) of one BraTS image sample. Various focuses (on different anatomical structures, e.g., tumor, brain, skull) are demonstrated by different channel-wise feature maps. 
Together with combined individual style code (sampling from a Gaussian distribution $\mathcal{N}(\bm{0},\bm{I})$), we only need to train one single ReMIC model to complete the multiple missing domains in the inputs.

In the image generation process (denoted as blue arrows in Fig.~\ref{fig:model}), our model samples style codes from a prior distribution and integrates with the content code to generate images in $N$ domains through generators $G_i$ ($1 \leq i \leq N$). The generator $G_i$ for each domain generates images in the corresponding domain from the domain-shared content code and the domain-specific style code: $G_i(c, s_i) = \tilde{x}_i$. 

Additionally, we extend the introduced image completion framework to a more practical scenario, i.e., tackling the missing data problem in image segmentation. Specifically, another branch of segmentation generator $G_S$ is added after content codes to generate the segmentation masks of the input images. Our underlying assumption is that the domain-shared content codes contain essential image structure information for the segmentation task. By simultaneously optimizing the generation loss and segmentation Dice loss (detailed in Section~\ref{sec:loss}), the model could adaptively learn how to generate missing images to improve the segmentation performance.

\subsection{Training Loss}
\label{sec:loss}
In the training of GAN models, the setting of losses is of paramount importance to the final generation results. Our loss functions contain the cycle-consistency loss of images and latent codes,
adversarial loss and reconstruction loss on the generated and input images.

\noindent \textbf{Image Consistency Loss:}
For each sample, the proposed model is able to extract a domain-shared content code and domain-specific style codes respectively from visible domains. Then by recombining the content and style codes, the domain generators are expected to recover the input images. The image consistency loss is defined to constrain the reconstructed images and real images as in the direction of ``Image $ \rightarrow$ Code $\rightarrow$ Image'' in Fig.~\ref{fig:model}.
\begin{equation}
    \begin{aligned}
    \mathcal{L}^{x_i}_{cyc} = \mathbb{E}_{x_i \sim p(x_i)} [\parallel  G_i(E^c(x_1, \cdots, x_N), E^s_i(x_i)) 
    &- x_i \parallel_1] 
    \end{aligned}
\vspace{-3pt}
\end{equation}
where $p(x_i)$ is the data distribution in domain $\mathbf{\chi}_i$ ($1 \leq i \leq N$). Here, we use $\mathcal{L}_1$ loss to strengthen anatomical-structure related generation.

\noindent \textbf{Latent Consistency Loss:}
The latent consistency loss constrains the learning of both content and style codes before decoding and after encoding in the direction of ``Code $\rightarrow$ Image $\rightarrow$ Code''.
\begin{equation}
    \begin{aligned}
        \mathcal{L}^{c}_{cyc} = &\mathbb{E}_{c \sim p(c), s_i \sim p(s_i)} [\parallel E^c(G_1(c, s_1), G_2(c, s_2), \cdots, G_n(c, s_N)) - c \parallel_1]
    \end{aligned}
\vspace{-3pt}
\end{equation}
\begin{equation}
    \begin{aligned}
        \mathcal{L}^{s_i}_{cyc} &= \mathbb{E}_{c \sim p(c), s_i \sim p(s_i)} [\parallel E^s_i(G_i(c, s_i)) - s_i \parallel_1] 
    \end{aligned}
\vspace{-3pt}
\end{equation}
where $p(s_i)$ is the prior distribution of style code: $\mathcal{N}(\bm{0}, \bm{I})$, $p(c)$ is given by $c = E^c(x_1, x_2, \cdots, x_N)$ and $x_i \sim p(x_i)$ ($1 \leq i \leq N$), i.e., the content code is sampled by firstly sampling images from data distribution. Specifically, taking BraTS data as an example, style distribution $p(s_i)$ contains various domain-specific characteristics in each domain, like varied image contrasts. Content distribution $p(c)$ contains various anatomy structure related features among different brain subjects as shown in Fig.~\ref{fig:content_brats}. 

{\small
\begin{figure}[t]
\centering
  \includegraphics[width=0.8\linewidth]{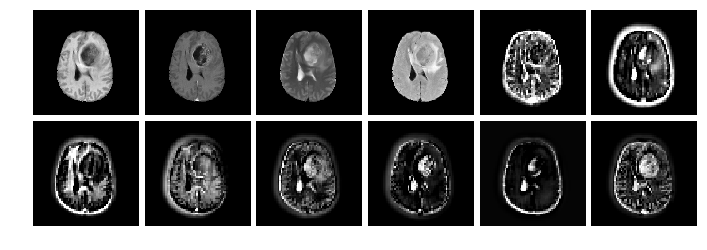}
  \caption{Content codes visualization in BraTS image generation. The first 4 images are ground truth modalities.
  }
\vspace{-0.2in}
\label{fig:content_brats}
\end{figure}}


\noindent \textbf{Adversarial Loss:}
The adversarial learning between generators and discriminators forces the data distribution of the generated images to be close to that of the real images for each domain.
\begin{equation}
    \begin{aligned}
    \mathcal{L}^{x_i}_{adv} = &\mathbb{E}_{c \sim p(c), s_i \sim p(s_i)} [\log (1 - D_i(G_i(c, s_i)))] + \mathbb{E}_{x_i \sim p(x_i)} [\log D_i(x_i)]
    \end{aligned}
\vspace{-3pt}
\end{equation}
where $D_i$ is the discriminator for domain $i$ to distinguish the generated images ${\tilde{x}_i}$ and real images $x_i \in \chi_i$.

\noindent \textbf{Reconstruction Loss:}
In addition to the feature-level consistency mentioned above to constrain the relationship between the generated images and real images in different domains, we also constrain the pixel-level similarity between generated images and ground truth images in the same domain during training stage, for accurately completing missing domains given visible images of the current subject or scene.
\begin{equation}
    \begin{aligned}
    \mathcal{L}^{x_i}_{rec} &= \mathbb{E}_{c \sim p(c), s_i \sim p(s_i)} [\parallel G_i(c, s_i) - x_i \parallel_1] 
    \end{aligned}
\vspace{-3pt}
\end{equation}

\noindent \textbf{Segmentation Loss:}
In the \(n\)-to-\(n\) image translation, the model learns a complementary representation of multiple domains, which can further facilitate the high-level tasks. For instance, extracted content code (containing the underlying anatomical structures) may benefit the segmentation of organs and lesions in medical image analysis, vice versa. 
Therefore, we train a multi-task network for both segmentation and generation. In the proposed framework, we construct a unified generation and segmentation model by adding a segmentation generator $G_S$ following the content code from the completed images as shown in Fig.~\ref{fig:model}. We utilize Dice loss~\cite{milletari2016vnet,salehi2017tversky} for accurate segmentation from multiple domain images 
\begin{equation}
    \begin{aligned}
    \mathcal{L}_{seg} = 1 - \frac{1}{L}{ \sum_{l=1}^{L} { \frac{ \sum_p {2 \hat{y}_p (l) y_p (l)} }{ \sum_p{\hat{y}_p (l)^2 + \sum_p y_p (l)^2}  } } },
    \end{aligned}\label{eq:seg_loss}
\vspace{-3pt}
\end{equation}
where $L$ is the total number of classes, $p$ is the spatial position index in the image, $\hat{y}(l)$ is the predicted segmentation probability map for class $l$ from $G_S$ and $y(l)$ is the ground truth segmentation mask for class $l$. The segmentation loss can be added into the total loss in Eq.~\ref{eq:total_loss} for an end-to-end joint learning optionally.

\noindent \textbf{Total Loss:}
The encoders, generators, discriminators (and segmentor) are jointly trained to optimize the total objective as follows
\begin{equation}
    \begin{aligned}
    & \min_{E^c, {E^s_1,\cdots,E^s_N,G_1,\cdots,G_N(, G_S)}} \max_{D_1,\cdots,D_N} \mathcal{L} (E^c, {E^s_1,\cdots,E^s_N}, {G_1,\cdots,G_N}, {D_1,\cdots,D_N}) \\
    &= \sum_{i=1}^{N} ( \lambda_{adv} \mathcal{L}^{x_i}_{adv} +  \lambda^x_{cyc} \mathcal{L}^{x_i}_{cyc}  + \lambda^s_{cyc} \mathcal{L}^{s_i}_{cyc} + \lambda_{rec} \mathcal{L}^{x_i}_{rec} )+ \lambda^c_{cyc} \mathcal{L}^{c}_{cyc} (+ \lambda_{seg} \mathcal{L}_{seg}),
    \end{aligned}\label{eq:total_loss}
\vspace{-3pt}
\end{equation}
where $\lambda_{adv}$, $\lambda^x_{cyc}$, $\lambda^c_{cyc}$, $\lambda^s_{cyc}$, $\lambda_{rec}$ and $\lambda_{seg}$ are hyper-parameters to balance the losses. Please note that the segmentation loss is included in the total training loss only when we train the unified generation and segmentation model for BraTS and ProstateX datasets.

\section{Experiments}

To validate the feasibility and generalization of the proposed model, we conduct experiments on two medical image datasets as well as a natural image dataset: BraTS, ProstateX, and RaFD. We firstly demonstrate the advantage of the proposed method in the \(n\)-to-\(n\) multi-domain image completion task given a random set of visible domains. Moreover, we illustrate that the proposed model (a variation with two branches of image translation and segmentation) provides an efficient solution to multi-domain segmentation with missing-domain inputs. 

\noindent \textbf{BraTS:} 
The Multimodal Brain Tumor Segmentation Challenge (BraTS) 2018~\cite{brats2014tmi,brats2017sci,brats2018identifying} provides multi-modal brain MRI with four modalities: a) native (T1),  b) post-contrast T1-weighted (T1Gd), c) T2-weighted (T2), and d) T2 Fluid Attenuated Inversion Recovery (FLAIR). Following CollaGAN~\cite{lee2019contrast}, 218 and 28 subjects are randomly selected for training and testing. A set of 2D slices is extracted from 3D volumes for four modalities respectively.
In total, the training and testing sets contain 40,148 and 5,340 images. We resize the images of size $240 \times 240$ to $256 \times 256$. Three categories are labeled for brain tumor segmentation, i.e., enhancing tumor (ET), tumor core (TC), and whole tumor (WT).

\noindent \textbf{ProstateX:}
The ProstateX dataset~\cite{litjens2014prostatex} contains multi-parametric prostate MR scans for 98 subjects. Each sample contains three modalities : 1) T2-weighted (T2), 2) Apparent Diffusion Coefficient (ADC), 3) high b-value DWI images (HighB). We randomly split it into 78 and 20 subjects for training and testing respectively. By extracting 2D slices from 3D volumes, the training and testing sets contain 3,540 and 840 images in total. Images of $384 \times 384$ are resized to $256 \times 256$. Prostate regions are manually labeled as the whole prostate (WP) by board-certificated radiologists.

\noindent \textbf{RaFD:}
The Radboud Faces Database (RaFD)~\cite{langner2010rafd} contains eight facial expressions collected from 67 participants: neutral, angry, contemptuous, disgusted, fearful, happy, sad, and surprised. Following StarGAN~\cite{choi2018stargan}, we adopt images from three camera angles (45$^{\circ}$, 90$^{\circ}$, 135$^{\circ}$) with three gaze directions (left, frontal, right), and obtain 4,824 images in total. 
The data is randomly split to training set of 54 participants (3,888 images) and testing set of 13 participants (936 images). 
We crop the image with the face in the center and then resize to $128 \times 128$. 

In all experiments, we set $\lambda_{adv} = 1, \lambda^x_{cyc} = 10, \lambda^c_{cyc} = 1, \lambda^s_{cyc} = 1, \lambda_{rec} = 20$, and $\lambda_{seg}=1$ if $\mathcal{L}_{seg}$ is included in Eq.~\ref{eq:total_loss}. 
The adversarial loss $\lambda_{adv}$ and consistency loss $\lambda^x_{cyc}, \lambda^c_{cyc}, \lambda^s_{cyc}$ follow the same loss weights choices as in~\cite{huang2018munit} which reported the necessity of the consistency losses in its ablative study. In the following, we will demonstrate ablative studies on the reconstruction and segmentation loss.


\section{Results}

\subsection{Results of Multi-Domain Image Completion}
For comparison purpose, we firstly assume there are only one missing domain for each data sample. In \textbf{training}, the one missing domain is randomly distributed among all the $N$ domains. During \textbf{testing}, at a time, we fix the one missing domain in inputs and evaluate the generation outputs only on that missing modality, whose results are demonstrated in one column (modality) of Table~\ref{tab:med_gen},~\ref{tab:rafd_gen}.
Multiple metrics are used to measure the similarity between the generated and teh target images, i.e., normalized root mean-squared error (NRMSE), mean structural similarity index (SSIM), and peak-signal-noise ratio (PSNR). 
We compare our results with previous methods on all three datasets. The results of the proposed method (``ReMIC''), ReMIC without reconstruction loss (``ReMIC w/o Recon'') are reported.

Moreover, we investigate a more practical scenario when there are more than one missing domains and show that our proposed method is capable to handle a general random \(n\)-to-\(n\) image completion. In this setting, we assume the set of missing domains in \textbf{training} data is randomly distributed, i.e. each training data has $k$ randomly selected visible domains where $k \geq 1$. During \textbf{testing}, we fix the number of visible domains $k$ ($k \in \{1,...,N-1\}$) while these $k$ available domains are also randomly distributed among $N$ domains. We evaluate all the $N$ generated images in outputs, showing results in all columns (modalities) of Table~\ref{tab:med_gen},~\ref{tab:rafd_gen}. ``ReMIC-Random($k=*$)'' denotes evaluation on the test set with $k$ random visible domains or $N-k$ random missing domains.
Note that by leveraging the unified content code and sampling the style code for each domain respectively, the proposed model could handle any number of missing domains, which is more general and flexible for the random \(k\)-to-\(n\) image completion as shown in Fig.~\ref{fig:motivation}(d). 
We compare our model with following methods:

\noindent \textbf{MUNIT}~\cite{huang2018munit} conducts \(1\)-to-\(1\) image translation between two domains through representational disentanglement as shown in Fig.~\ref{fig:motivation}(a). In RaFD experiments, we train and test MUNIT models between any pair of two domains. Without loss of generality, we use ``neural'' image to generate all the other domains by following StarGAN setting, and ``angry'' image is used to generate ``neural'' image. In BraTS , the typical modality ``T1'' is used to generate other domains while ``T1'' is generated from ``T1Gd''. Similarly, ``T2'' is used to generate other domains in ProstateX while it is generated from ``ADC''. 

\noindent \textbf{StarGAN}~\cite{choi2018stargan} adopts a mask vector to generate image in the specified target domain. In this way, different target domains could be generated from one source domain in multiple inference passes. This is actually a \(1\)-to-\(n\) image translation as in Fig.~\ref{fig:motivation}(b). Since only one domain can be used as input in StarGAN, we use the same domain pair match as MUNIT, following the same setting in~\cite{choi2018stargan}.

\begin{figure}[t]
\hfill
\subfigure[Single missing modality.]{
    \includegraphics[width=0.5\linewidth]{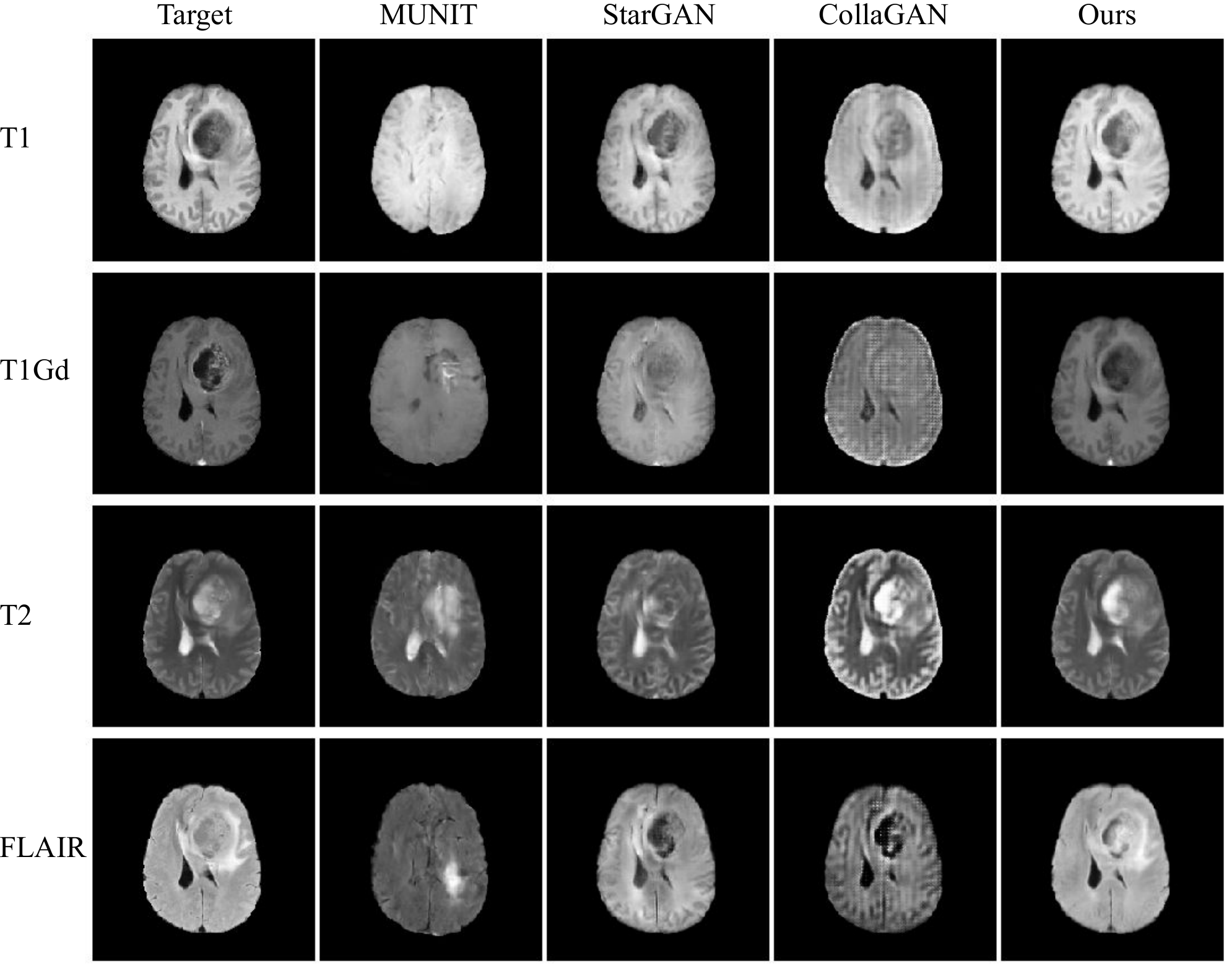}
    \label{fig:brats_gen}
    }
\hfill
\subfigure[Multiple missing modalities.]{
    \includegraphics[width=0.43\linewidth]{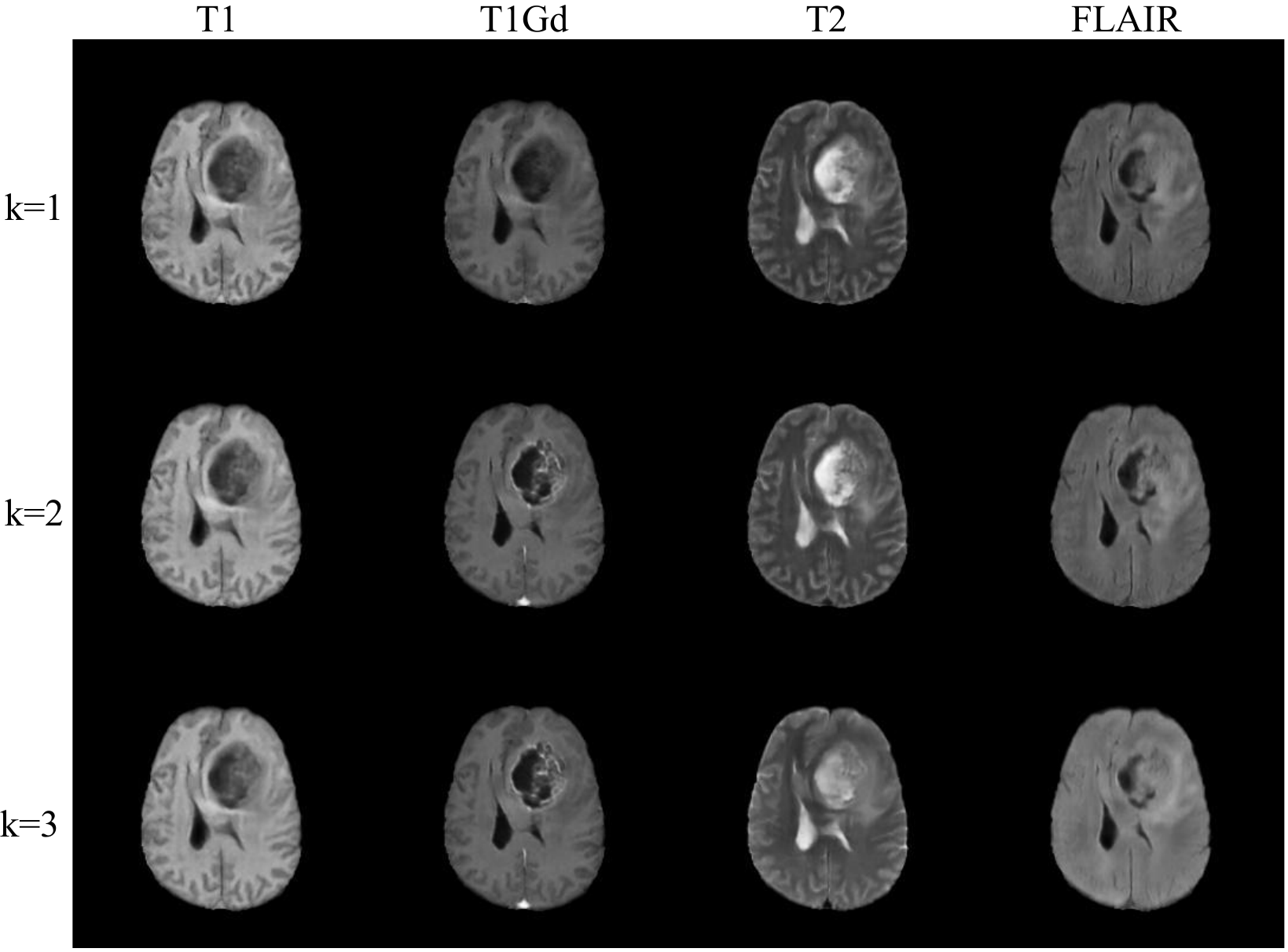}
    \label{fig:brats_gen_rand}
    }
\hfill
\caption{(a) BraTS image generation results with a single missing modality. Rows: 4 modalities. Columns: compared methods. (b) BraTS image generation results with multiple missing modalities (in columns). Ground truth image for each modality is shown in (a). Rows: the first $k$ domains (from left to right) are given in inputs ($1 \leq k \leq 3$).}
\end{figure}

\noindent \textbf{CollaGAN}~\cite{lee2019collagan,lee2019contrast} carries out the \(n\)-to-\(1\) image translation in Fig.~\ref{fig:motivation}(c), where multiple source domains collaboratively generate one target domain which is assumed missing in inputs. But it does not deal with multiple missing domains. In CollaGAN experiments, we use the same domain generation setting as ours, i.e., fix one missing domain in inputs and generate from all the other domains.




\begin{figure}[t]
\hfill
\subfigure[Single missing modality.]{
    \includegraphics[width=0.52\linewidth]{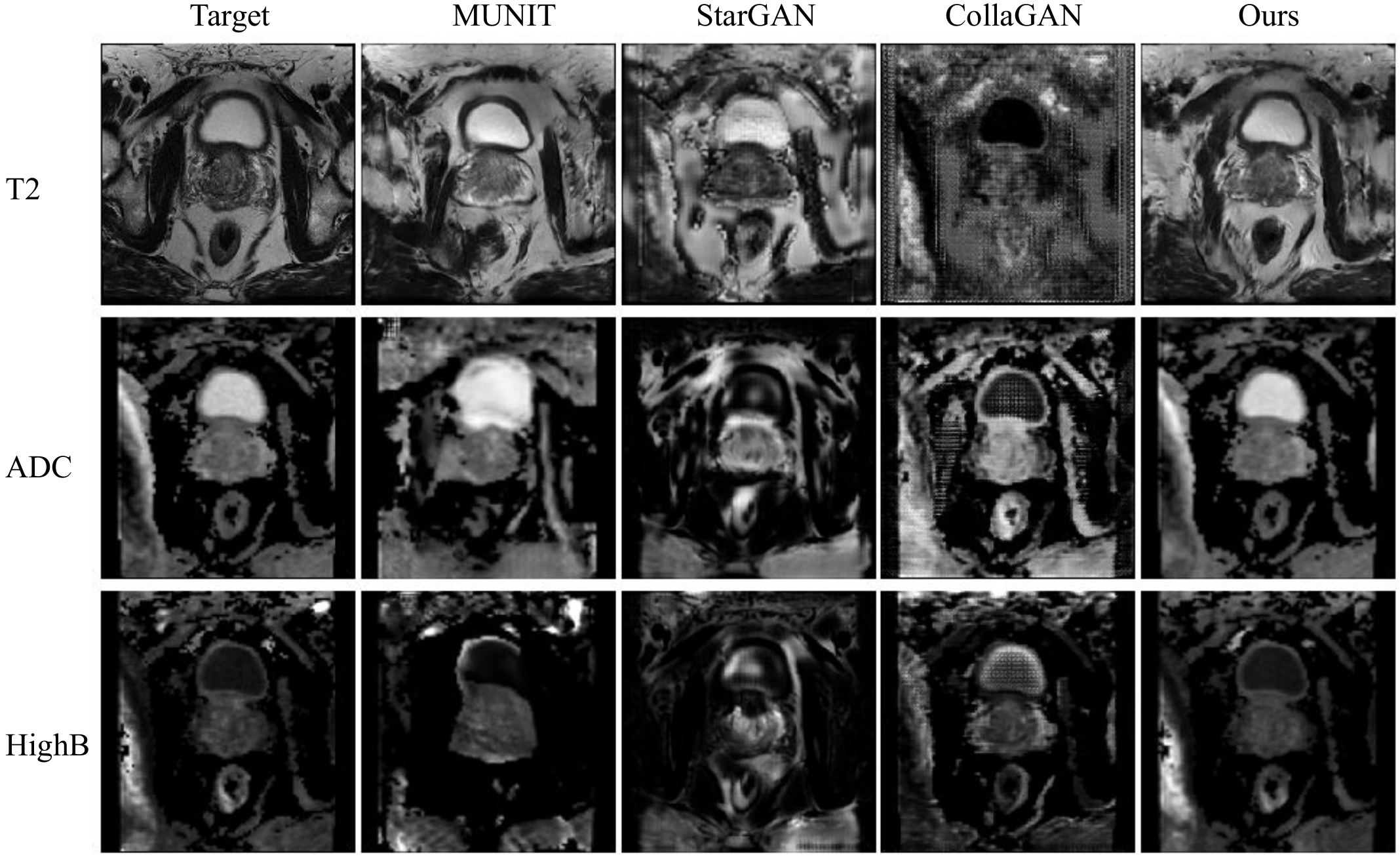}
    \label{fig:prostate_gen}
    }
\hfill
\subfigure[Multiple missing modalities.]{
    \includegraphics[width=0.4\linewidth]{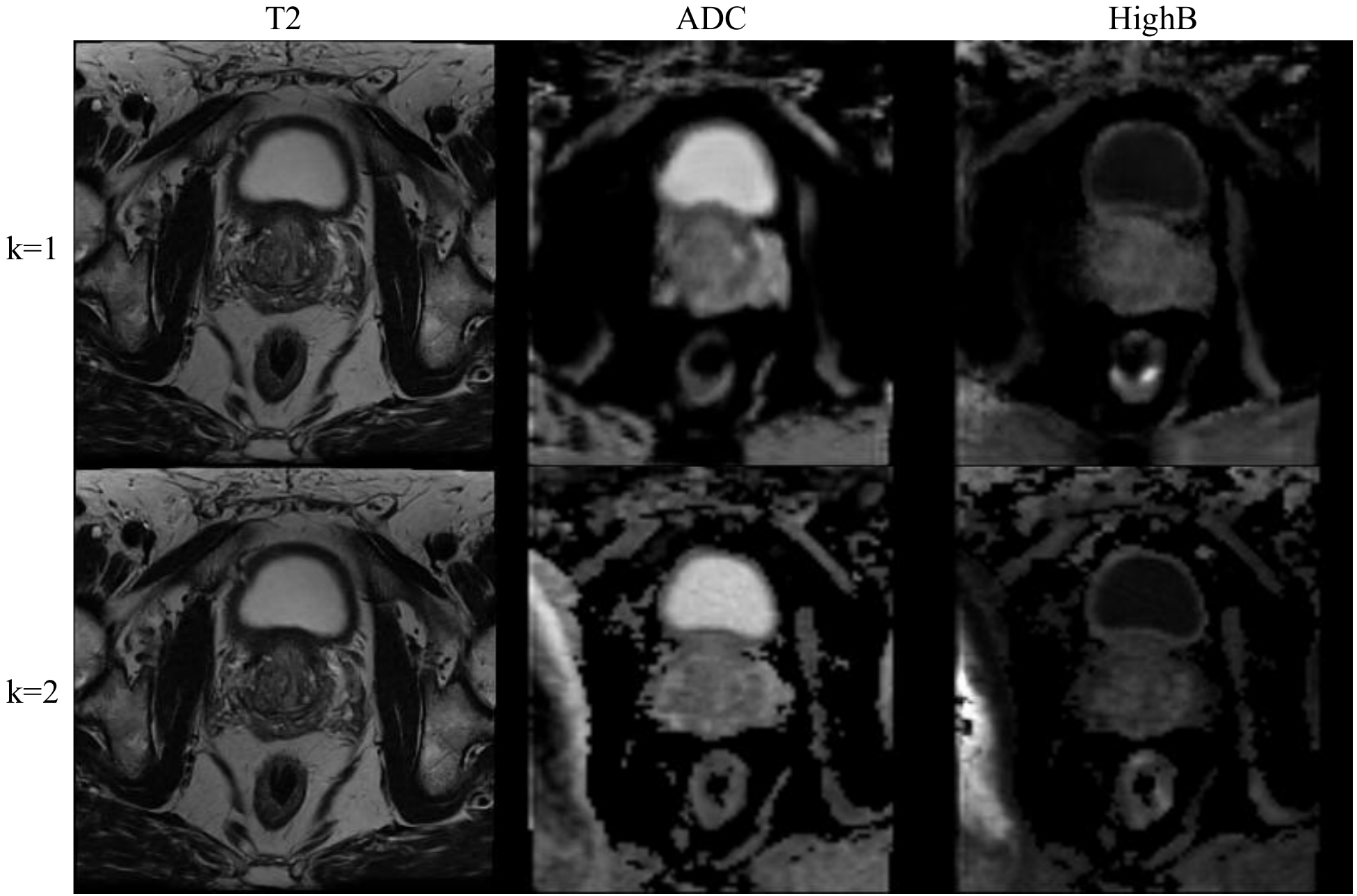}
    \label{fig:prostate_gen_rand}
    }
\hfill
\caption{(a) ProstateX image generation results with a single missing modality. Rows: 3 modalities. Columns: compared methods. (b) ProstateX image generation results with multiple missing modalities (in columns). Ground truth image for each modality is shown in (a). Rows: the first $k$ domains (from left to right) are given in inputs ($1 \leq k \leq 2$).}
\end{figure}


\begin{table*}[t]
\centering
\caption{BraTS and Prostate multi-domain image completion results.}
\label{tab:med_gen}
\centering
\resizebox{1.0\textwidth}{!}{
\begin{tabular}{l|l|l|l|l}
\multicolumn{5}{c}{(a) BraTS} \\
\Xhline{2\arrayrulewidth}

\multicolumn{1}{c|}{Methods} & \multicolumn{1}{c|}{T1} & \multicolumn{1}{c|}{T1Gd} & \multicolumn{1}{c|}{T2} & \multicolumn{1}{c}{FLAIR}
\\ \cline{2-5}
\multicolumn{1}{c|}{}  & {NRMSE($\downarrow$) / SSIM($\uparrow$) / PSNR($\uparrow$)} & {NRMSE($\downarrow$) / SSIM($\uparrow$) / PSNR($\uparrow$)} & {NRMSE($\downarrow$) / SSIM($\uparrow$) / PSNR($\uparrow$)} & {NRMSE($\downarrow$) / SSIM($\uparrow$) / PSNR($\uparrow$)}
\\ 
\Xhline{2\arrayrulewidth}
{MUNIT~\cite{huang2018munit}} & {0.3709 / 0.9076 / 23.2385} & {0.2827 / 0.9221 / 27.3836} & {0.4073 / 0.8757 / 22.8936} & {0.4576 / 0.8702 / 21.5568} \\
{StarGAN~\cite{choi2018stargan}} & {0.3233 / 0.9282 / 24.2840} & {0.2718 / 0.9367 / 27.6901} & {0.5002 / 0.8464 / 21.3614} & {0.4642 / 0.8855 / 22.0483} \\ 
{CollaGAN~\cite{lee2019collagan}} & {0.4800 / 0.8954 / 21.2803} & {0.4910 / 0.8706 / 22.9042} & {0.5310 / 0.8886 / 21.2163} & {0.4231 / 0.8635 / 22.4188} \\  
\Xhline{2\arrayrulewidth}
{ReMIC w/o Recon} & {0.3366 / 0.9401 / 24.5787} & {0.2398 / 0.9435 / 28.8571} & {0.3865 / 0.9011 / 23.4876} & {0.3650 / 0.8978 / 23.5918} \\ 
{ReMIC} & {\textbf{0.2008} / \textbf{0.9618} / \textbf{28.5508}} & {\textbf{0.2375} / \textbf{0.9521} / \textbf{29.1628}} & {\textbf{0.2481} / \textbf{0.9457} / \textbf{27.4829}} & {\textbf{0.2469} / \textbf{0.9367} / \textbf{27.1540}} \\
\Xhline{2\arrayrulewidth}
{ReMIC-Random(k=1)} & {0.2263 / 0.9603 / 27.5198} & {0.2118 / 0.9600 / 30.5945} & {0.2566 / 0.9475 / 27.7646} & {0.2742 / 0.9399 / 26.8257} \\
{ReMIC-Random(k=2)} & {0.1665 / 0.9751 / 30.8579} & {0.1697 / 0.9730 / 32.7615} & {0.1992 / 0.9659 / 30.3789} & {0.2027 / 0.9591 / 29.7351} \\
{ReMIC-Random(k=3)} & {0.1274 / 0.9836 / 33.2458} & {0.1405 / 0.9812 / 34.3967} & {0.1511  / 0.9788 / 32.6743} & {0.1586 / 0.9724 / 31.8967} \\
\Xhline{2\arrayrulewidth}
\end{tabular}}

\centering
\resizebox{0.8\textwidth}{!}{
\begin{tabular}{l|l|l|l}
\multicolumn{4}{c}{(b) ProstateX} \\
\Xhline{2\arrayrulewidth}
\multicolumn{1}{c|}{Methods} & \multicolumn{1}{c|}{T2} & \multicolumn{1}{c|}{ADC} & \multicolumn{1}{c}{HighB}
\\ \cline{2-4}
\multicolumn{1}{c|}{}  & NRMSE($\downarrow$) / SSIM($\uparrow$) / PSNR($\uparrow$) & NRMSE($\downarrow$) / SSIM($\uparrow$) / PSNR($\uparrow$) & NRMSE($\downarrow$) / SSIM($\uparrow$) / PSNR($\uparrow$) \\ 
\Xhline{2\arrayrulewidth}
MUNIT~\cite{huang2018munit} & {0.6904 / 0.4428 / 15.6308} & {0.9208 / 0.4297 / 13.8983} & {0.9325 / 0.5383 / 16.9616} \\ 
StarGAN~\cite{choi2018stargan} & {0.6638 / 0.4229 / 15.9468} & {0.9157 / 0.3665 / 13.8014} & {0.9188 / 0.4350 / 17.1168} \\
CollaGAN~\cite{lee2019collagan} & {0.8070 / 0.2667 / {14.2640}} & {0.7621 / 0.4875 / 15.4242} & {0.7722 / 0.6824 / 18.6481} \\  
\Xhline{2\arrayrulewidth}
ReMIC w/o Recon & {0.8567 / 0.3330 / 13.6738} & {0.7289 / 0.5377 / 15.7083} & {0.8469 / 0.7818 / 17.8987} \\
ReMIC & {\textbf{0.4908} / \textbf{0.5427} / \textbf{18.6200}} & {\textbf{0.2179} / \textbf{0.9232} / \textbf{26.6150}} & {\textbf{0.3894} / \textbf{0.9150} / \textbf{24.7927}} \\ 
\Xhline{2\arrayrulewidth}
{ReMIC-Random(k=1)}  & {0.3786 / 0.6569 / 22.5314} & {0.2959 / 0.8256 / 26.9485} & {0.4091 / 0.8439 / 27.7499} \\
{ReMIC-Random(k=2)}  & {0.2340 / 0.8166 / 27.0598} & {0.1224 / 0.9664 / 33.2475} & {0.1958 / 0.9587 / 34.4775} \\
\Xhline{2\arrayrulewidth}
\end{tabular}}
\vspace{-0.1in}
\end{table*}

\noindent \textbf{Results of medical image generation:}
Fig.~\ref{fig:brats_gen} and Fig.~\ref{fig:prostate_gen} show the results of image completion (modalities in rows) on BraTS and ProstateX data in comparison to others~\cite{huang2018munit,choi2018stargan,lee2019collagan} (methods in columns). Each cell illustrates the generated image when the current modality is missing in inputs. The corresponding quantitative results averaged across all testing data are shown in Table~\ref{tab:med_gen}. In comparison, our model generates better results in meaningful details, e.g., a more accurate outstanding tumor region in BraTS and prostate regions are better-preserved in ProstateX. This is achieved by learning a better content code through factorized latent space in our method, which is essential in preserving the anatomical structures in medical images.
Furthermore, we illustrate the generation results when multiple modalities are missing in BraTS and ProstateX dataset. We show the results in the rows of Fig.~\ref{fig:brats_gen_rand} and Fig.~\ref{fig:prostate_gen_rand}, where images are generated when only the first $k$ modalities (from left to right) are given in the inputs ($1 \leq k \leq N-1$). The averaged quantitative results for random $k$-to-$n$ image generation are denoted as ``ReMIC-Random($k=*$)'' in Table~\ref{tab:med_gen}.

{\small
\begin{figure}[t]
\centering
  \includegraphics[width=0.8\linewidth]{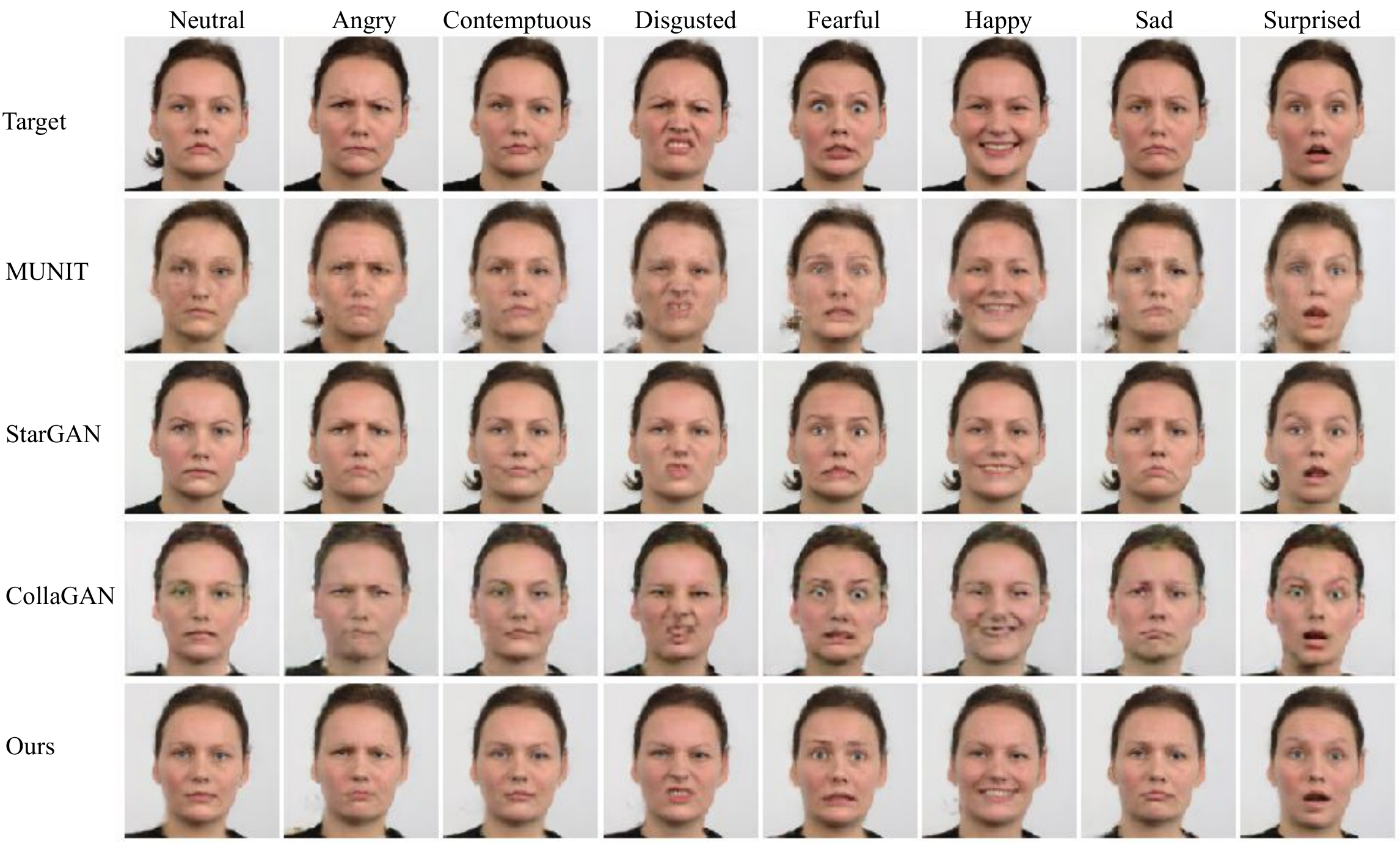}
  \caption{RaFD image generation results with a single missing modality. Columns: 8 facial expressions. Rows: compared methods.}
\label{fig:rafd_gen}
\vspace{-0.1in}
\end{figure}}

{\small
\begin{figure}[t]
\centering
  \includegraphics[width=0.8\linewidth]{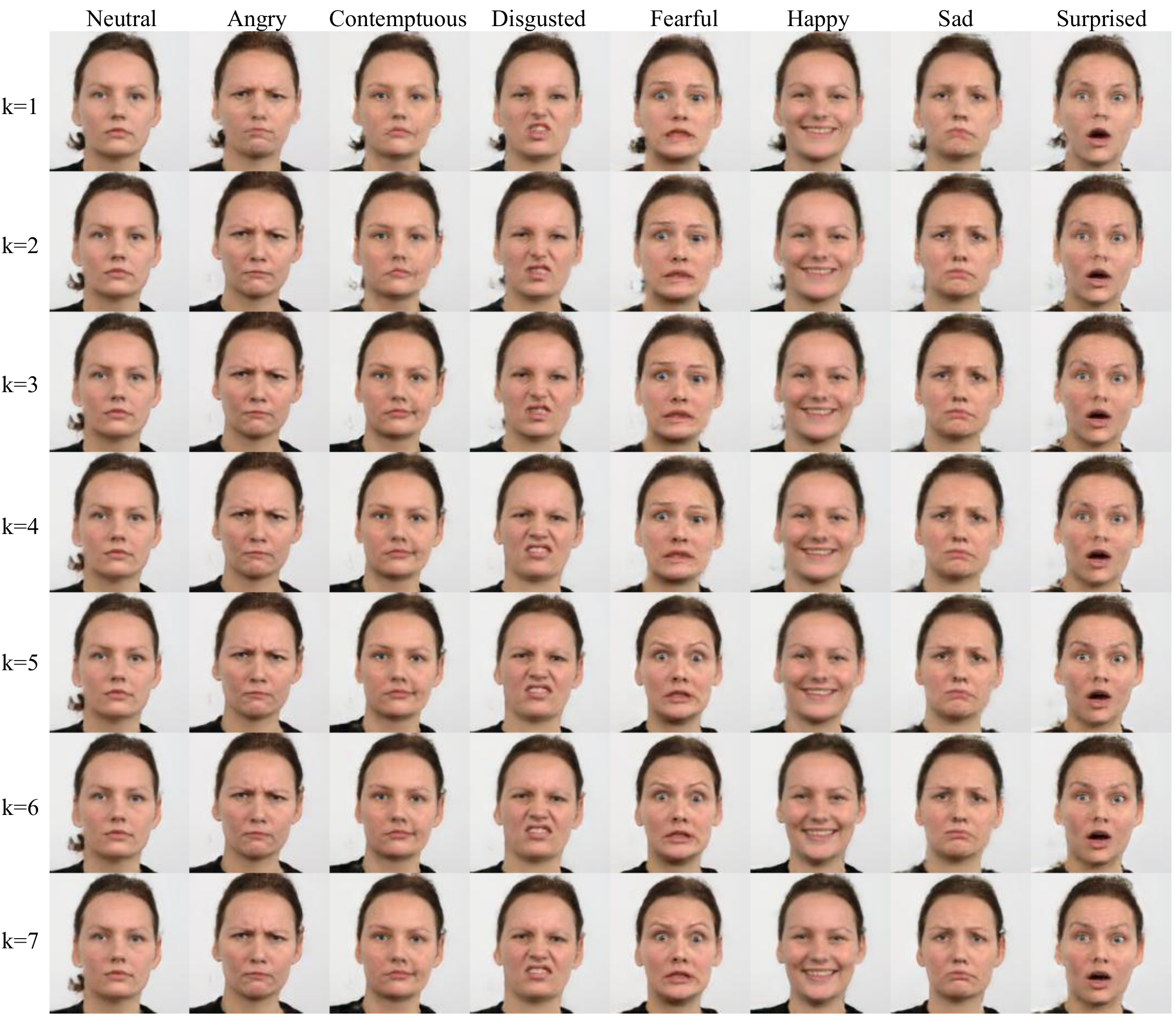}
  \caption{RaFD image generation results with multiple missing modalities (in columns). Ground truth image for each modality is shown in ``Target'' row of Fig.~\ref{fig:rafd_gen}. Rows: the first $k$ domains (from left to right) are given in inputs ($1 \leq k \leq 7$).
  }
\label{fig:rafd_gen_rand}
\vspace{-0.1in}
\end{figure}}

\begin{table*}[t]
\centering
\caption{RaFD multi-domain image completion results.}
\label{tab:rafd_gen}
\resizebox{1.\textwidth}{!}{
\begin{tabular}{l|l|l|l|l}
\Xhline{2\arrayrulewidth}
\multicolumn{1}{c|}{Methods} & \multicolumn{1}{c|}{Neutral} & \multicolumn{1}{c|}{Angry} & \multicolumn{1}{c|}{Contemptuous} & \multicolumn{1}{c}{Disgusted}
\\ \cline{2-5}
\multicolumn{1}{c|}{}  & {NRMSE($\downarrow$) / SSIM($\uparrow$) / PSNR($\uparrow$)} & {NRMSE($\downarrow$) / SSIM($\uparrow$) / PSNR($\uparrow$)} & {NRMSE($\downarrow$) / SSIM($\uparrow$) / PSNR($\uparrow$)} & {NRMSE($\downarrow$) / SSIM($\uparrow$) / PSNR($\uparrow$)}
\\ 
\Xhline{2\arrayrulewidth}
{MUNIT~\cite{huang2018munit}} & {0.1589 / 0.8177 / 19.8469} & {0.1637 / 0.8156 / 19.7303} & {0.1518 / 0.8319 / 20.2793} & {0.1563 / 0.8114 / 19.9362} \\ 
{StarGAN~\cite{choi2018stargan}} & {0.1726 / 0.8206 / 19.2725} & {0.1722 / 0.8245 / 19.4336} & {0.1459 / 0.8506 / 20.7605} & {0.1556 / 0.8243 / 20.0036} \\ 
{CollaGAN~\cite{lee2019collagan}} & {0.1867 / 0.7934 / 18.3691} & {0.1761 / 0.7736 / 18.8678} & {0.1856 / 0.7928 / 18.4040} & {0.1823 / 0.7812 / 18.5160} \\  
\Xhline{2\arrayrulewidth}
{ReMIC w/o Recon} & {\textbf{0.1215} / 0.8776 / \textbf{22.2963}} & {0.1335 / 0.8556 / 21.4615} & {\textbf{0.1192} / \textbf{0.8740} / \textbf{22.4073}} & {0.1206 / 0.8559 / 22.1819} \\ 
{ReMIC} & {0.1225 / \textbf{0.8794} / 22.2679} & {\textbf{0.1290} / \textbf{0.8598} / \textbf{21.7570}} & {0.1217 / 0.8725 / 22.2414} & {\textbf{0.1177} / \textbf{0.8668} / \textbf{22.4135}} \\
\Xhline{2\arrayrulewidth}
{ReMIC-Random(k=1)} & {0.1496 / 0.8317 / 20.7821} & {0.1413 / 0.8368 / 21.5096} & {0.1407 / 0.8348 / 21.2486} & {0.1394 / 0.8352 / 21.4443} \\
{ReMIC-Random(k=4)} & {0.0990 / 0.9014 / 24.7746} & {0.0988 / 0.8964 / 24.8327} & {0.0913 / 0.9048 / 25.2826} & {0.0969 / 0.8934 / 24.8231} \\
{ReMIC-Random(k=7)}  & {0.0756 / 0.9280 / 26.6861} & {0.0679 / 0.9332 / 27.4557} & {0.0665 / 0.9346 / 27.5942} & {0.0675 / 0.9308 / 27.3955} \\
\Xhline{2\arrayrulewidth}
\multicolumn{1}{c|}{Methods} & \multicolumn{1}{c|}{Fearful} & \multicolumn{1}{c|}{Happy} & \multicolumn{1}{c|}{Sad} & \multicolumn{1}{c}{Surprised}
\\ \cline{2-5}
\multicolumn{1}{c|}{}  & {NRMSE($\downarrow$) / SSIM($\uparrow$) / PSNR($\uparrow$)} & {NRMSE($\downarrow$) / SSIM($\uparrow$) / PSNR($\uparrow$)} & {NRMSE($\downarrow$) / SSIM($\uparrow$) / PSNR($\uparrow$)} & {NRMSE($\downarrow$) / SSIM($\uparrow$) / PSNR($\uparrow$)} \\ 
\Xhline{2\arrayrulewidth}
{MUNIT~\cite{huang2018munit}} & {0.1714 / 0.7792 / 19.1714} & {0.1623 / 0.8073 / 19.7709} & {0.1677 / 0.7998 / 19.3867} & {0.1694 / 0.7884 / 19.3867} \\ 
{StarGAN~\cite{choi2018stargan}} & {0.1685 / 0.7943 / 19.3516} & {0.1522 / 0.8288 / 20.4397} & {0.1620 / 0.8227 / 19.7368} & {0.1634 / 0.7974 / 19.6744} \\ 
{CollaGAN~\cite{lee2019collagan}} & {0.1907 / 0.7442 / 18.1518} & {0.1829 / 0.7601 / 18.5503} & {0.1783 / 0.7766 / 18.7450} & {0.1888 / 0.7495 / 18.2169} \\  
\Xhline{2\arrayrulewidth}
{ReMIC w/o Recon} & {0.1321 / 0.8384 / 21.4604} & {0.1399 / 0.8332 / 20.9334} & {\textbf{0.1284} / \textbf{0.8597} / \textbf{21.7430}} & {0.1333 / 0.8347 / 21.3782} \\ 
{ReMIC} & {\textbf{0.1316} / \textbf{0.8395} / \textbf{21.5295}} & {\textbf{0.1383} / \textbf{0.8406}/ \textbf{21.0465}} & {0.1301 / 0.8581 / 21.6384} & {\textbf{0.1276} / \textbf{0.8484} / \textbf{21.7793}} \\ 
\Xhline{2\arrayrulewidth}
{ReMIC-Random(k=1)} & {0.1479 / 0.8132 / 21.0039} & {0.1567 / 0.8121 / 20.3798} & {0.1491 / 0.8244 / 20.6888} & {0.1434 / 0.8218 / 21.2411} \\
{ReMIC-Random(k=4)} & {0.1043 / 0.8769 / 24.2623} & {0.1065 / 0.8852 / 23.9813} & {0.0960 / 0.8971 / 24.9114} & {0.1022 / 0.8835 / 24.2613} \\
{ReMIC-Random(k=7)}  & {0.0769 / 0.9209 / 26.5362} & {0.0794 / 0.9200 / 26.1515} & {0.0729 / 0.9291 / 26.8993} & {0.0735 / 0.9248 / 26.7651} \\
\Xhline{2\arrayrulewidth}
\end{tabular}}
\vspace{-0.1in}
\end{table*}

\begin{table}[t]
\centering
\caption{Missing-domain segmentation. Dice scores are reported.}
\label{tab:seg}
\resizebox{0.7\textwidth}{!}{
\begin{tabular}{l|l|l|l|l|l|l|l}
\Xhline{2\arrayrulewidth}
\multicolumn{1}{c|}{Methods} & \multicolumn{4}{c|}{BraTS} & \multicolumn{3}{c}{ProstateX} \\
\cline{2-8}
\multicolumn{1}{c|}{} & \multicolumn{1}{c|}{T1} & \multicolumn{1}{c|}{T1Gd} & \multicolumn{1}{c|}{T2} & \multicolumn{1}{c|}{FLAIR} & \multicolumn{1}{c|}{T2} & \multicolumn{1}{c|}{ADC} & \multicolumn{1}{c}{HighB} \\
\Xhline{2\arrayrulewidth}
{Oracle+All} & \multicolumn{4}{c|}{0.822} & \multicolumn{3}{c}{0.908} \\ 
\Xhline{2\arrayrulewidth}
{Oracel+Zero padding} & {0.651} & {0.473} & {0.707} & {0.454} & {0.528} & {0.243} & {0.775} \\ 
{Oracle+Average imputation} & {0.763} & {0.596} & {0.756} & {0.671} & {0.221} & {0.692} & {0.685} \\ 
{Oracle+Nearest neighbor} & {0.769} & {0.540} & {0.724} & {0.606} & {0.759} & {0.850} & {0.854} \\
\cline{1-8}
{Oracle+MUNIT} & {0.783} & {0.537} & {0.782} & {0.492} & {0.783} & {0.708} & {0.858} \\ 
{Oracle+StarGAN} & {0.799} & {0.553} & {0.746} & {0.613} & {0.632} & {0.653} & {0.832} \\ 
{Oracle+CollaGAN} & {0.753} & {0.564} & {0.798} & {0.674} & {0.472} & {0.760} & {0.842} \\

{Oracle+ReMIC} & {0.789} & {0.655} & {0.805} & {0.765} & {0.871} & {0.898} & {0.891} \\
\Xhline{2\arrayrulewidth}
{ReMIC+Seg} & {0.806} & {0.674} & {0.822} & {0.771} & {\textbf{0.872}} & {\textbf{0.909}} & {\textbf{0.905}} \\
{ReMIC+Joint} & {\textbf{0.828}} & {\textbf{0.693}} & {\textbf{0.828}} & {\textbf{0.791}} & {0.867} & {0.904} & {0.904} \\ 
\Xhline{2\arrayrulewidth}
\end{tabular}}
\vspace{-0.1in}
\end{table}

\noindent \textbf{Results of facial expression image generation:}
Fig.~\ref{fig:rafd_gen} shows the result of facial expression image completion for RaFD dataset. In each column, we show the target and generated images of each domain (facial expression), where we assume the current target domain is missing in the inputs at a time and needs to be generated using the rest $N-1$ available domains. 
Compared with MUNIT and StarGAN results, our method could generate missing images with a better quality, especially in generating details like teeth, mouth and eyes. This benefits from that our method can incorporate complementary information from multiple available domains, while MUNIT and StarGAN can adopt only one domain as input. For example, in the generation of ``happy'' and ``disgusted'' expressions, either MUNIT nor StarGAN could generate a good teeth and mouth region, since their source domain ``neutral'' does not contain the teeth. Compared with CollaGAN, our method could generate images with a better content due to the explicit disentangled representational learning in feature level instead of the implicit cycle-consistency constraints only in pixel level.
Moreover, Fig.~\ref{fig:rafd_gen_rand} shows the results of multiple missing domains. Each row shows the generated images in each of 8 domains, when the first $k$ domains (from left to right) are given in inputs ($1 \leq k \leq 7$). The superior performance could also be observed in the NRMSE, and SSIM and PSNR evaluation metrics averaged across all testing samples as reported in Table~\ref{tab:rafd_gen} with all the eight expression domains.

\subsection{Results of Missing-Domain Segmentation }
Based on the missing-domain image completion, we demonstrate that our proposed method could go beyond image generation to solve the missing-domain image segmentation. Specifically, our model learns factorized representations by disentangling latent space, which could be efficiently leveraged for high-level segmentation task. As shown in Fig.~\ref{fig:model}, a segmentation branch is added using the learned content code to generate segmentation prediction.  We evaluate the segmentation performance with dice coefficient on both BraTS and ProstateX datasets as shown in Table~\ref{tab:seg}. Please note that we show the average dice coefficient across three categories for BraTS dataset: enhancing tumor (ET), tumor core (TC), and whole tumor (WT). (details of per-category results in supplementary.)

We train a fully supervised 2D U-shaped segmentation network (a U-Net variation~\cite{ronneberger2015unet}) without missing images as the ``Oracle''.  ``Oracle+*'' means that the results are computed by testing the missing images generated or imputed from the ``*'' method with the pretrained ``Oracle'' model. ``All'' represents the full testing set without any missing domains. ``ReMIC+Seg'' stands for using separate content encoders for image generation and segmentation tasks in our proposed unified framework, while ``ReMIC+Joint'' indicates sharing the weights of content encoder for the two tasks. For the results on both datasets, our proposed unified framework with joint training of image generation and segmentation could achieve the best segmentation performance in comparison to other imputation or generation methods. Moreover, it even obtains comparable results as ``Oracle'' model when some modalities are missing. This indicates that the learned content codes indeed embed and extract efficient anatomical structures for image representation.

In our experiments, we choose the widely used U-shaped segmentation network~\cite{ronneberger2015unet} as the backbone for segmentation generator $G_S$. Here, we focus on showing how the proposed method could benefit the segmentation when missing domains exist and the segmentation backbone is fixed. But our method can also be easily generalized to other segmentation models with a similar methodology.

\section{Conclusion}
In this work, we propose a general framework for multi-domain image completion, given that one or more input domains are missing. The proposed model learns shared content and domain-specific style encoding across multiple domains. 
We show the proposed image completion approach can be well generalized to both natural and medical images.
Our framework is further extended to a unified image generation and segmentation framework to tackle a practical problem of missing-domain segmentation. Experiments on three datasets demonstrate the proposed method consistently achieves better performance than several previous approaches on both multi-domain image completion and segmentation with random missing domains.

%
%
\bibliographystyle{splncs04}
\bibliography{ReMIC_ref}
\appendix
\section{Implementation Details}
Here, we describe the implementation details of our method. We will also open source all the source codes and models upon the acceptance of this work.

\subsection{Hyperparameters}
In our algorithm, we use the Adam optimizer~\cite{kingma2014adam} with $\beta_1 = 0.5, \beta_2 = 0.999$. The learning rate is 0.0001. We set the loss weights in the total loss (Equation 7 in main text) as $\lambda_{adv} = 1, \lambda^x_{cyc} = 10, \lambda^c_{cyc} = 1, \lambda^s_{cyc} = 1, \lambda_{rec} = 20$, and $\lambda_{seg} = 1$ in the unified model for image completion and segmentation.
For comparison purpose, we train the model with batch size 1 and 100,000 iterations for image generation task, and compare the results across MUNIT~\cite{huang2018munit}, StarGAN~\cite{choi2018stargan}, CollaGAN~\cite{lee2019collagan}, and ours ReMIC in all the three datasets. 
In ReMIC, we set the dimension of the style code as 8 for comparison purpose with MUNIT. For image generation during testing, we use a fixed style code of 0.5 in each dimension for both MUNIT and ReMIC to compute quantitative results.

\subsection{Network Architectures}
The network structure of ReMIC is developed on the backbone of MUNIT model~\cite{huang2018munit}. We describe the details of each module here.

\noindent \textbf{Unified Content Encoder}: 
consists of a down-sampling module and residual blocks to extract contextual knowledge from all available domain images in inputs. The down-sampling module contains a $7 \times 7$ convolutional block with stride 1 and 64 filters, and two $4 \times 4$ convolutional blocks with stride 2 and, 128 and 256 filters respectively. The convolutional layers downsample the input to features maps of size $W/4 \times H/4 \times 256$, where $W$ and $H$ are the width and height of input image. Next, there are four residual blocks, each of which contains two $3 \times 3$ convolutional blocks with 256 filters and stride 1. We apply Instance Normalization (IN)~\cite{ulyanov2017instancenorm} after all the convolutional layers. Note that the proposed unified content encoder accepts images of all domains as input (missing domains are filled up with zeros padding in the initialization), and learns a universe content code complementarily and collaboratively, which are different from MUNIT.

\noindent \textbf{Style Encoder}:
contains a similar down-sampling module and several residual blocks, which is followed by a global average pooling layer and a fully connected layer to learn the verteorized style code. The down-sampling module is developped using the same structure as that in the unified content encoder above, and two more $4 \times 4$ convolutional blocks with stride 2 and 256 filters are followed. The final fully connected layer generates style code as a 8-dim vector. There is no IN applied to the style encoders to keep the original feature means and variances with style information~\cite{huang2017adain}.

\noindent \textbf{Generator}:
includes four residual blocks, each of which contains two $3 \times 3$ convolutional blocks with 256 filters and stride 1. Two nearest-neighbor upsampling layers and a $5 \times 5$ convolutional block with stride 1 and, 128 and 64 filters respectively are followed to up-sample content codes back to the original image size. Finally, there is a a $7 \times 7$ convolutional block with stride 1 to output the reconstructed image. In order to incorporate the style code in the generation process, the Adaptive Instance Normalization (AdaIN)~\cite{huang2017adain} is applied to each residual block as follows~\cite{huang2018munit}:
\begin{align}
    \textit{AdaIN}(z, \gamma, \beta) = \gamma \Big( \frac{z-\mu(z)}{\sigma(z)} \Big) + \beta
\end{align}
where $z$ is the activation from the last convolutional layer. $\mu(z)$ and $\sigma(z)$ are the channel-wise mean and standard deviations of the activation. $\gamma$ and $\beta$ are the affine parameters in the AdaIN layers that are generated from style codes via a multi-layer perceptron (MLP). In this way, the input style code controls the generated style information through the affine transformation in the AdaIN layers in all generators~\cite{huang2017adain}.

\noindent \textbf{Discriminator}:
includes four $4 \times 4$ convolutional blocks with stride 2 and, 64, 128, 256, and 512 filters in sequence. The Leaky ReLU activation with slope 0.2 is applied after convolutional layers. A multi-scale discriminator~\cite{wang2018multiscale} is used to include the results at three different scales together. In adversarial training, we adopt LSGAN objective~\cite{mao2017lsgan} as the adversarial loss to learn to generate realistic images.

\noindent \textbf{Segmentor}:
We adopt a segmentation net with a U-Net shape~\cite{ronneberger2015unet}. In order to build a joint model with the image generation modules, we build a variant of U-Net, that is, the downsampling part shares the same structure as the content encoder aforementioned while the upsampling part has the same layers as the generator as described above. Similar to the original U-Net~\cite{ronneberger2015unet}, we also adopt the skip connections between the downsampling and upsampling layers in our segmentation module.

\section{Extended Ablative Study and Results for Multi-domain Image Completion}
\begin{table*}[t]
\centering
\caption{Extended results of multi-domain image completion for BraTS dataset}
\label{tab:brats_gen_full}
\resizebox{1.0\textwidth}{!}{
\begin{tabular}{l|l|l}
\Xhline{2\arrayrulewidth}
\multicolumn{1}{c|}{Methods} & \multicolumn{1}{c|}{T1} & \multicolumn{1}{c}{T1Gd} \\ 
\cline{2-3}
\multicolumn{1}{c|}{}  & {MAE($\downarrow$) / NRMSE($\downarrow$) / PSNR($\uparrow$) / SSIM($\uparrow$)} & {MAE($\downarrow$) / NRMSE($\downarrow$) / PSNR($\uparrow$) / SSIM($\uparrow$)} \\ 
\cline{1-3}
{ReMIC} & {0.0187 / {0.2008} / {28.5508} / {0.9618}} & {0.0153 / {0.2375} / {29.1628} / {0.9521}} \\
{ReMIC+Multi-Sample} & {\textbf{0.0180} / \textbf{0.1942} / \textbf{28.8354} / \textbf{0.9634}} & {\textbf{0.0127} / \textbf{0.2070} / \textbf{30.2444} / \textbf{0.9555}} \\ 
\cline{1-3}
{ReMIC+Seg} & {0.0195 / 0.2033 / 28.5679 / {0.9597}} & {{0.0142} / {0.2285} / {29.2134} / {0.9468}} \\ 
{ReMIC+Joint} & {{0.0214} / {0.2128} / 27.9944 / 0.9568} & {0.0140 / 0.2251 / 29.3624 / 0.9484} \\ 
\Xhline{2\arrayrulewidth}
\multicolumn{1}{c|}{Methods} & \multicolumn{1}{c|}{T2} & \multicolumn{1}{c}{FLAIR}
\\ \cline{2-3}
\multicolumn{1}{c|}{}  & {MAE / NRMSE / PSNR / SSIM} & {MAE / NRMSE / PSNR / SSIM} \\ 
\cline{1-3}
{ReMIC} & {0.0190 / \textbf{0.2481} / 27.4829 /{0.9457} } & {0.0198 / {0.2469} / 27.1540 / {0.9367}} \\
{ReMIC+Multi-Sample} & {{0.0195} / {0.2493} / \textbf{27.5168} / \textbf{0.9463}} & {\textbf{0.0192} / \textbf{0.2456} / \textbf{27.3598} / \textbf{0.9385}} \\ 
\cline{1-3}
{ReMIC+Seg} & {\textbf{0.0193} / 0.2525 / 27.2864 / 0.9431} & {0.0206 / 0.2553 / 26.9191 / 0.9333} \\ 
{ReMIC+Joint} & {{0.0197} / 0.2596 / 26.9954 / 0.9429} & {0.0220 / 0.2651 / 26.5068 / 0.9302} \\ 
\Xhline{2\arrayrulewidth}
\end{tabular}}
\end{table*}

\begin{table*}[t]
\centering
\caption{Extended results of multi-domain image completion for ProstateX dataset}
\label{tab:pt_gen_full}
\resizebox{1.0\textwidth}{!}{
\begin{tabular}{l|l|l}
\Xhline{2\arrayrulewidth}
\multicolumn{1}{c|}{{Methods}} & \multicolumn{1}{c|}{T2} & \multicolumn{1}{c}{ADC} \\ 
\cline{2-3}
\multicolumn{1}{c|}{}  & {MAE($\downarrow$) / NRMSE($\downarrow$) / PSNR($\uparrow$) / SSIM($\uparrow$)} & {MAE($\downarrow$) / NRMSE($\downarrow$) / PSNR($\uparrow$) / SSIM($\uparrow$)} \\ 
\cline{1-3}
ReMIC & {0.0840 / 0.4908 / 18.6200 / 0.5427} & {0.0253 / 0.2179 / 26.6150 / 0.9232} \\ 
ReMIC+Multi-Sample & {0.0810 / \textbf{0.4742} / 18.8986 / \textbf{0.5493}} & {\textbf{0.0250} / \textbf{0.2171} / \textbf{26.7024} / \textbf{0.9263}} \\ 
\cline{1-3}
ReMIC+Seg & {\textbf{0.0871} / 0.5024 / 18.4236 / 0.5336} & {0.0272 / 0.2322 / 26.0828 / 0.9107} \\
ReMIC+Joint & {0.0881 / 0.5071 / 18.3206 / 0.5353} & {0.0288 / 0.2403 / 25.8024 / 0.9064} \\ 
\Xhline{2\arrayrulewidth}
\multicolumn{1}{c|}{{Methods}} & \multicolumn{1}{c|}{HighB}
\\ \cline{2-2}
\multicolumn{1}{c|}{} & {MAE($\downarrow$) / NRMSE($\downarrow$) / PSNR($\uparrow$) / SSIM($\uparrow$)} \\
\cline{1-2}
ReMIC & {\textbf{0.0254} / \textbf{0.3894} / 24.7927 / \textbf{0.9150}} \\ 
ReMIC+Multi-Sample & {0.0268 / 0.3945 / \textbf{24.8066} / 0.9116} \\ 
\cline{1-2}
ReMIC+Seg & {0.0272 / 0.4110 / 24.3277 / 0.9061} \\
ReMIC+Joint & {0.0286 / 0.4359 / 23.8270 / 0.9006} \\ 
\Xhline{2\arrayrulewidth}
\end{tabular}}
\end{table*}

In this section, we conduct more ablative studies in multi-domain image completion with multi-sample learning and multi-task learning with the unified model of image generation and segmentation. More quantitative results are demonstrated in Table~\ref{tab:brats_gen_full} and Table~\ref{tab:pt_gen_full}, which are the corresponding extended tables for Table 1 in the main text. Please note that in addition to the metrics of NRMSE, SSIM and PSNR, we also add the MAE metric to measure the difference between generated images and the ground truth.

\subsection{Multi-sample learning}
\label{sec:multi-sample}
Based on the proposed model as shown in Fig. 3 in the main text, we further propose a training strategy when multiple samples are inputted at one time to facilitate learning disentangled representations. Specifically, based on the assumption of partially shared latent space, we assume that the factorized latent code can represent the corresponding content and style information in the input image. Therefore, by exchanging the style codes from two independent samples in all available domains, it should be able to reconstruct the original input images by recombining the original content and the new style codes from the other sample. Based on this idea, we build a comprehensive model with cross-sample training between two samples. Similarly as the framework in Fig. 3 in main text, the image and latent consistency loss and image reconstruction loss are also constrained through the encoding and decoding procedure. The results of multi-sample learning are shown in Table~\ref{tab:brats_gen_full} and Table~\ref{tab:pt_gen_full} denoted as ``ReMIC+Multi-Sample''.

\subsection{Multi-task learning}

For the jointly trained model of image completion and segmentation, the generated images are also evaluated using the same metrics as shown in Table~\ref{tab:brats_gen_full} and Table~\ref{tab:pt_gen_full}. Similarly to Table 3 in the main text, ``ReMIC+Seg'' stands for using separate content encoders for image generation and segmentation tasks in our proposed unified framework, while ``ReMIC+Joint'' indicates sharing the weights of content encoder for both two tasks. The results indicate that adding segmentation branch does not bring an obvious benefit for image generation. This is because the segmentation sub-module mainly focuses on the tumor region which takes up only a small part among the whole slice image. Besides, we use dice loss as the segmentation training objective which might not be consistent with the metrics used to evaluate generated image quality, which mainly emphasize the whole-slice pixel-level similarity.

\subsection{Random multi-domain image completion}

As described in Section 5.1 of the main text,  we investigate a more practical scenario when there are more than one missing domains and show that our proposed method is capable to handle a general random \(n\)-to-\(n\) image completion. In this setting, we assume the set of missing domains in \textbf{training} data is randomly distributed, i.e. each training sample has $k$ randomly selected domains where $k$ is at least 1. During \textbf{testing}, we evaluate the model with different number of existing domains $k$ ($k \in \{1,...,N-1\}$) while these $k$ available domains are in the order of from domain 1 to domains $N$. 

\begin{figure*}[h]
\centering
  \includegraphics[width=0.8\linewidth]{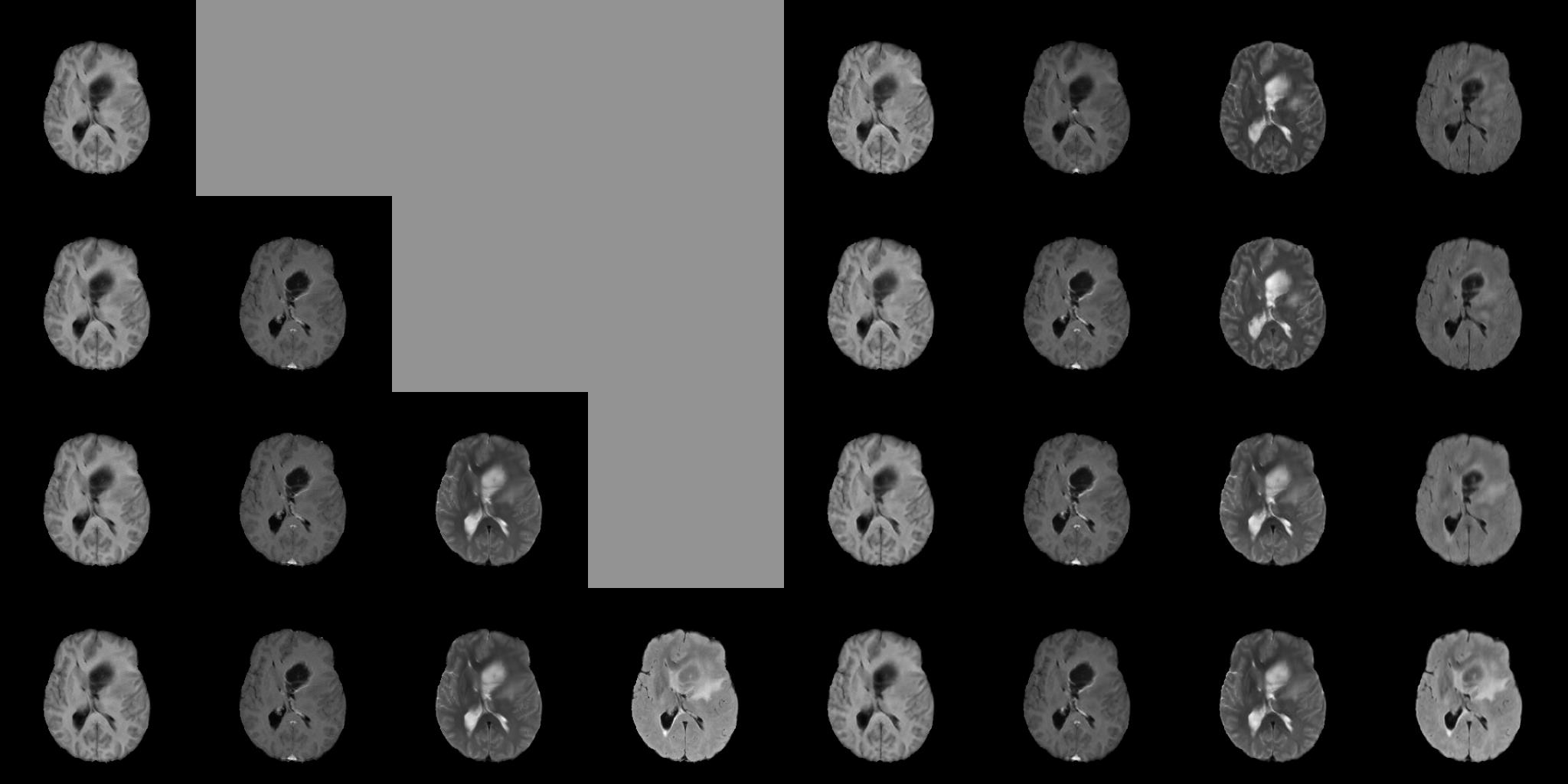}
  \includegraphics[width=0.8\linewidth]{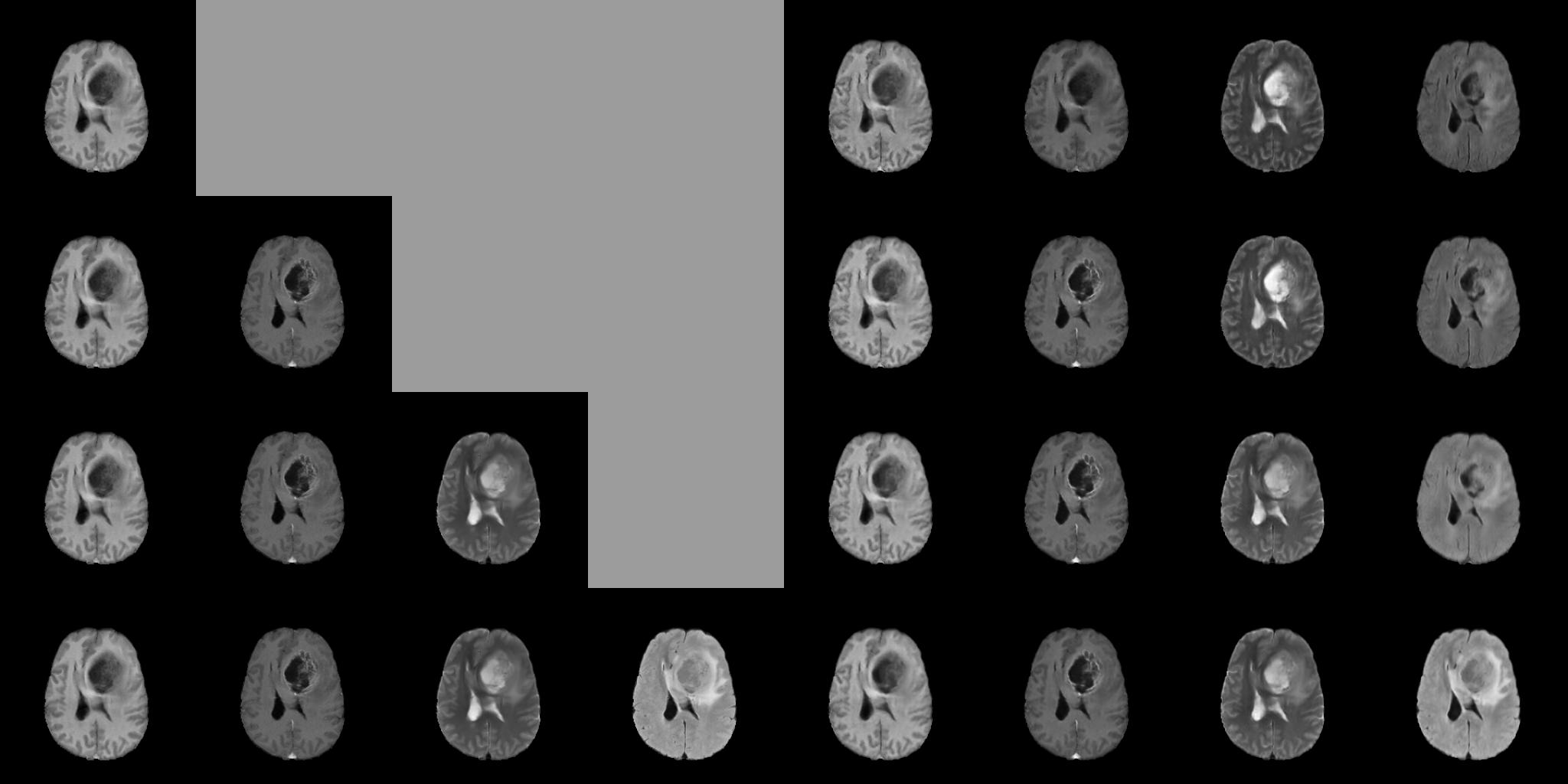}
  \includegraphics[width=0.8\linewidth]{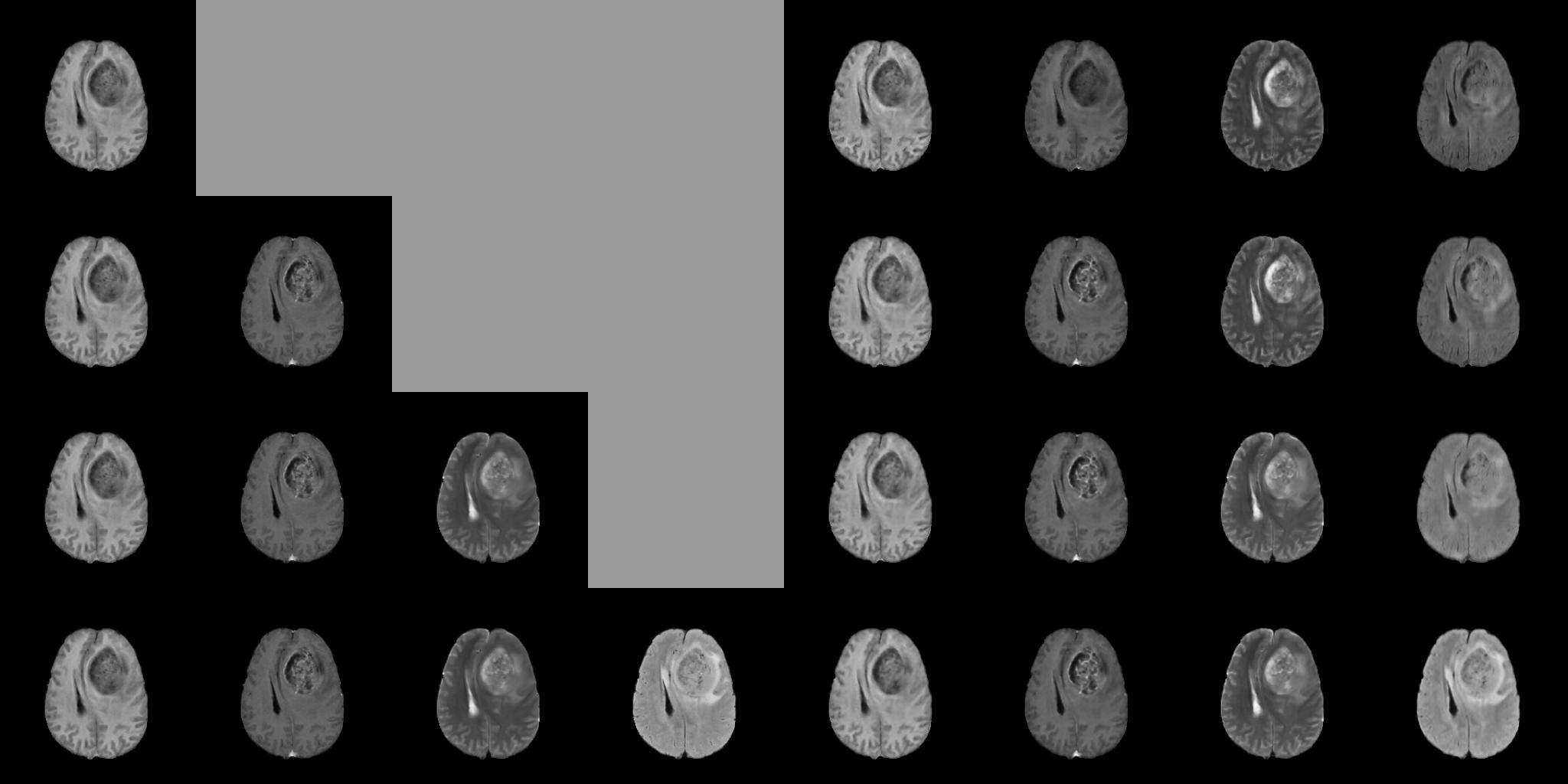}
  \caption{Random multi-domain image completion results of three testing samples in BraTS. The completed images (right) are generated from partial existing images in inputs (left). The number of inputted visible domains is in the range of $[1, N]$ where $N=4$ is the number of all domains. Rows: every 4 rows show results for one testing sample. Columns: 4 image modalities of T1, T1Gd, T2, and FLAIR.}
\label{fig:brats_n2n}
\end{figure*}

\begin{figure*}[h]
\centering
  \includegraphics[width=0.8\linewidth]{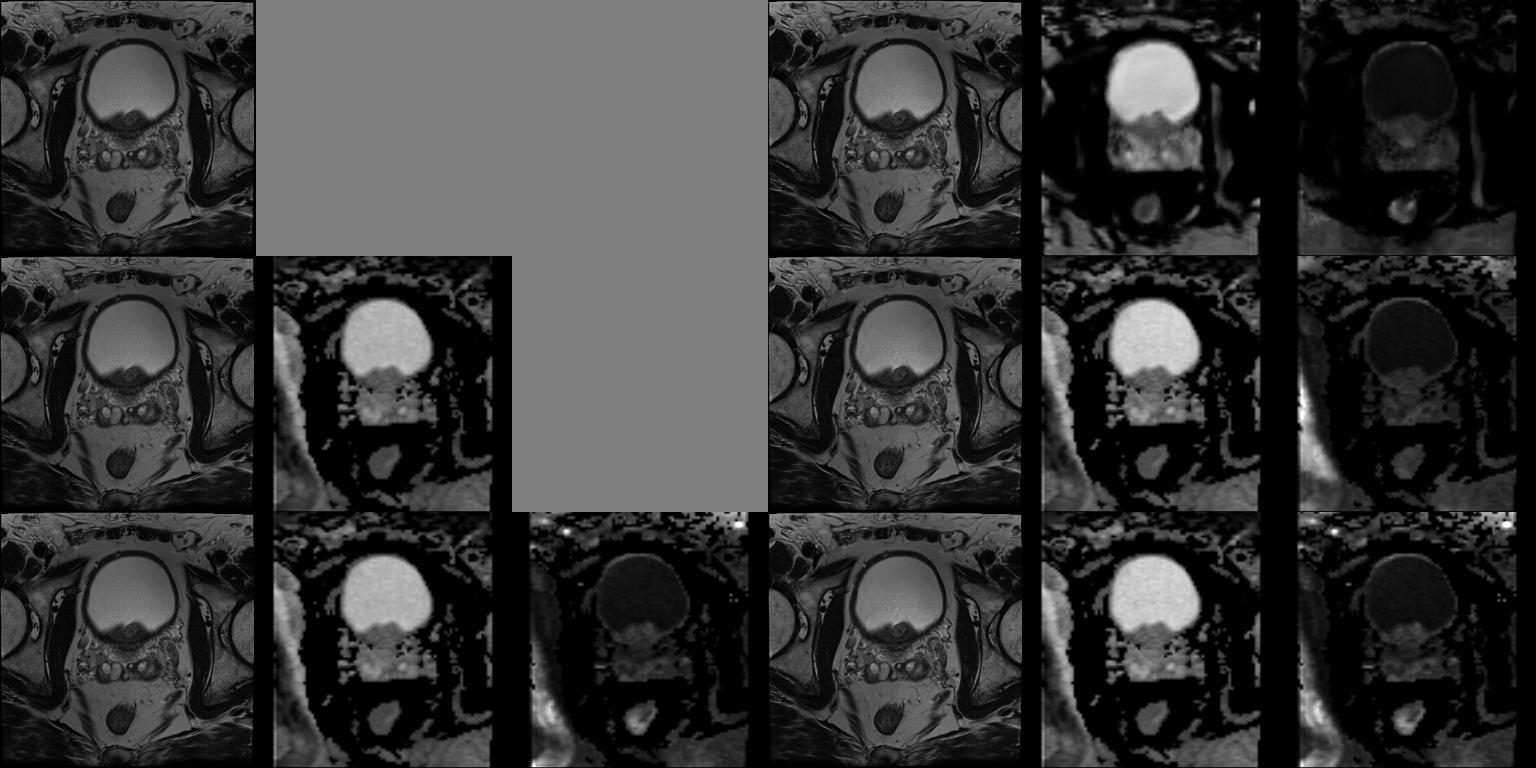}
  \includegraphics[width=0.8\linewidth]{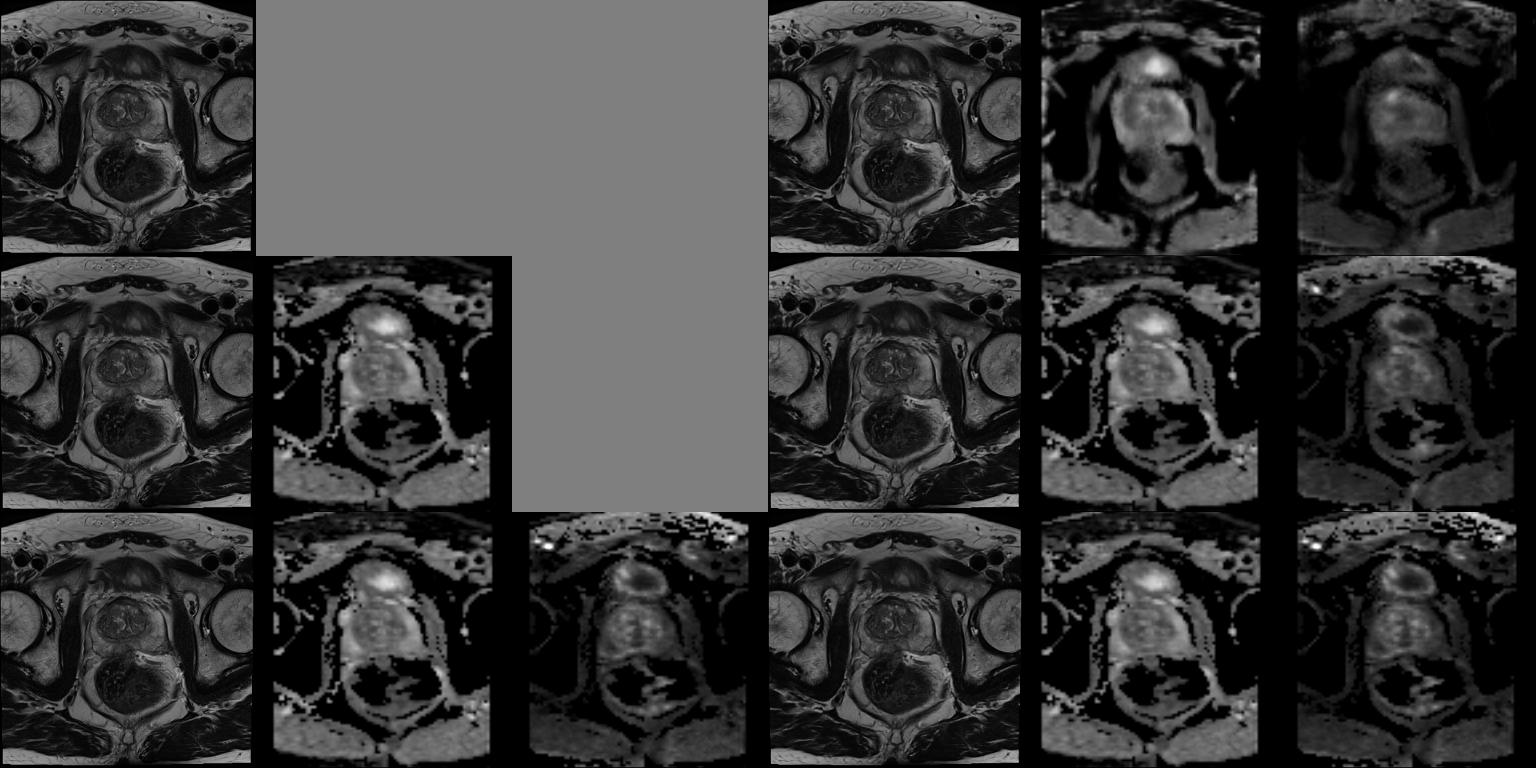}
  \includegraphics[width=0.8\linewidth]{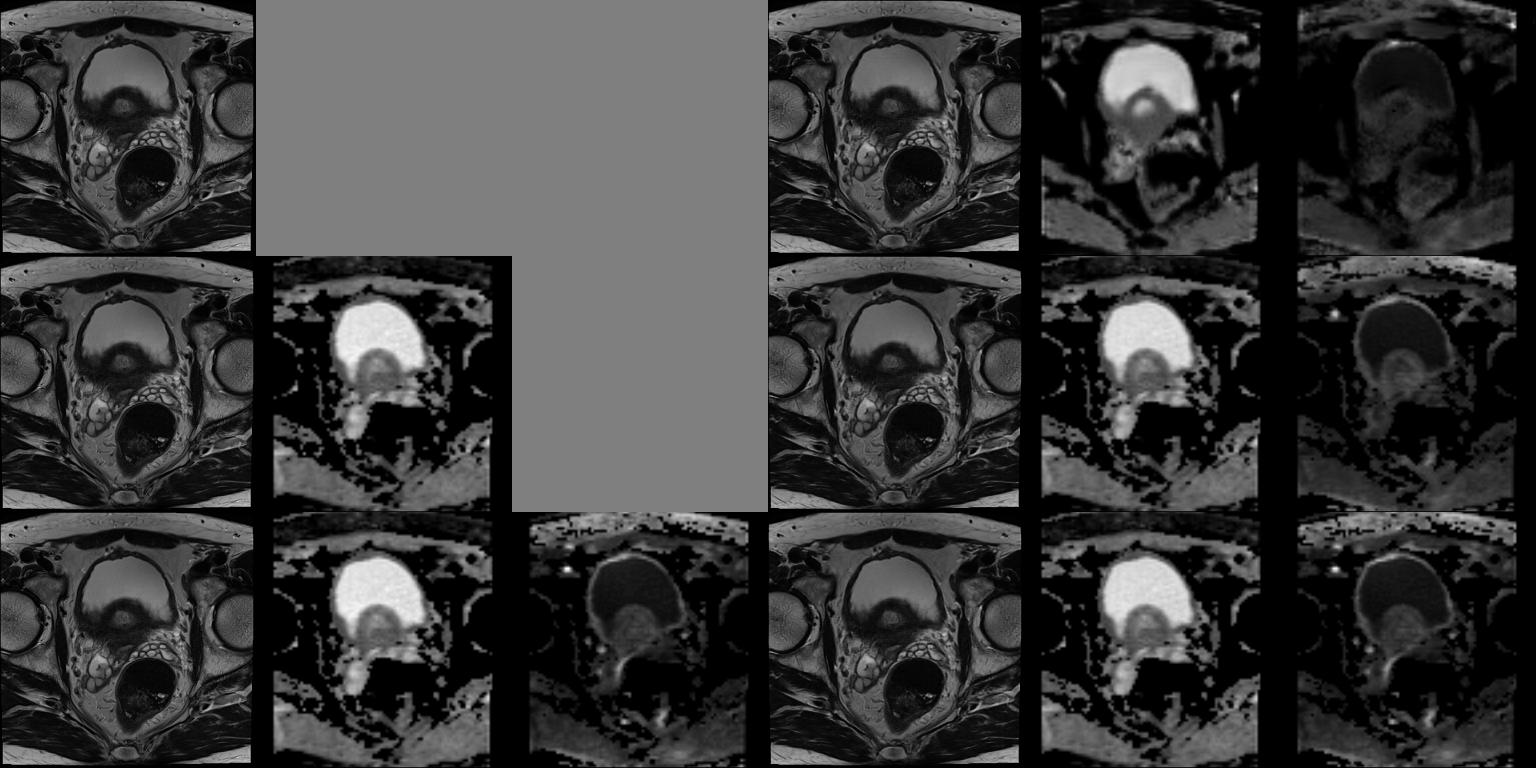}
  \caption{Random multi-domain image completion results of three testing samples in ProstateX. The completed images (right) are generated from partial existing images in inputs (left). The number of inputted visible domains is in the range of $[1, N]$ where $N=3$ is the number of all domains. Rows: every 3 rows show results for one testing sample. Columns: 3 image modalities of T2, ADC, and HighB.}
\label{fig:prostate_n2n}
\end{figure*}

\begin{figure*}[h]
\centering
  \includegraphics[width=0.6\linewidth]{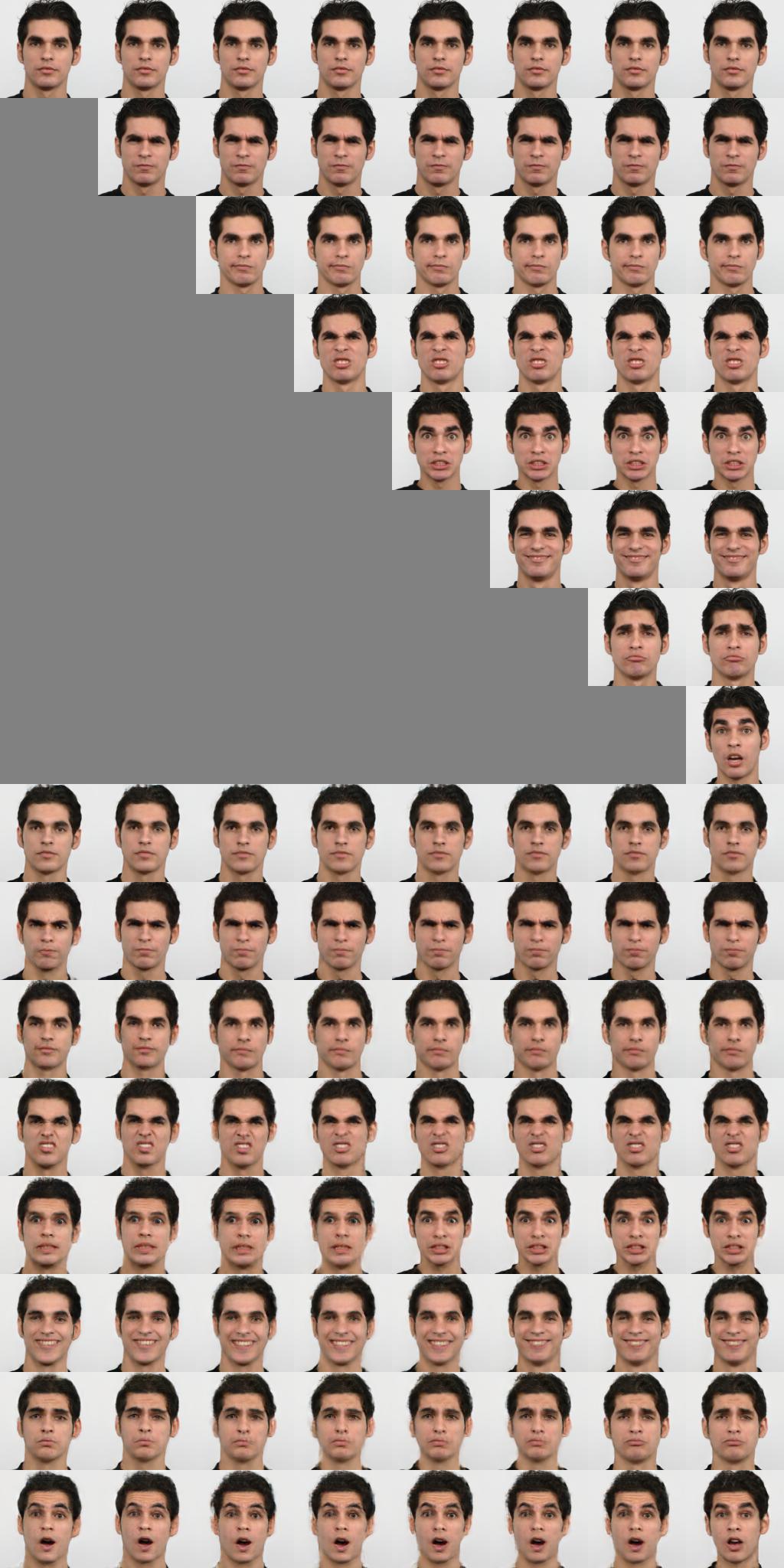}
  \caption{Random multi-domain image completion results of three testing samples in RaFD. The completed images (bottom) are generated from partial existing images in inputs (top). The number of inputted visible domains is in the range of $[1, N]$ where $N=8$ is the number of all domains. Rows: 8 image domains of ``neutral'', ``angry'', ``contemptuous'', ``disgusted'', ``fearful'', ``happy'', ``sad'', ``surprised''.
  }
\label{fig:face_n2n_1}
\end{figure*}

\begin{figure*}[h]
\centering
  \includegraphics[width=0.6\linewidth]{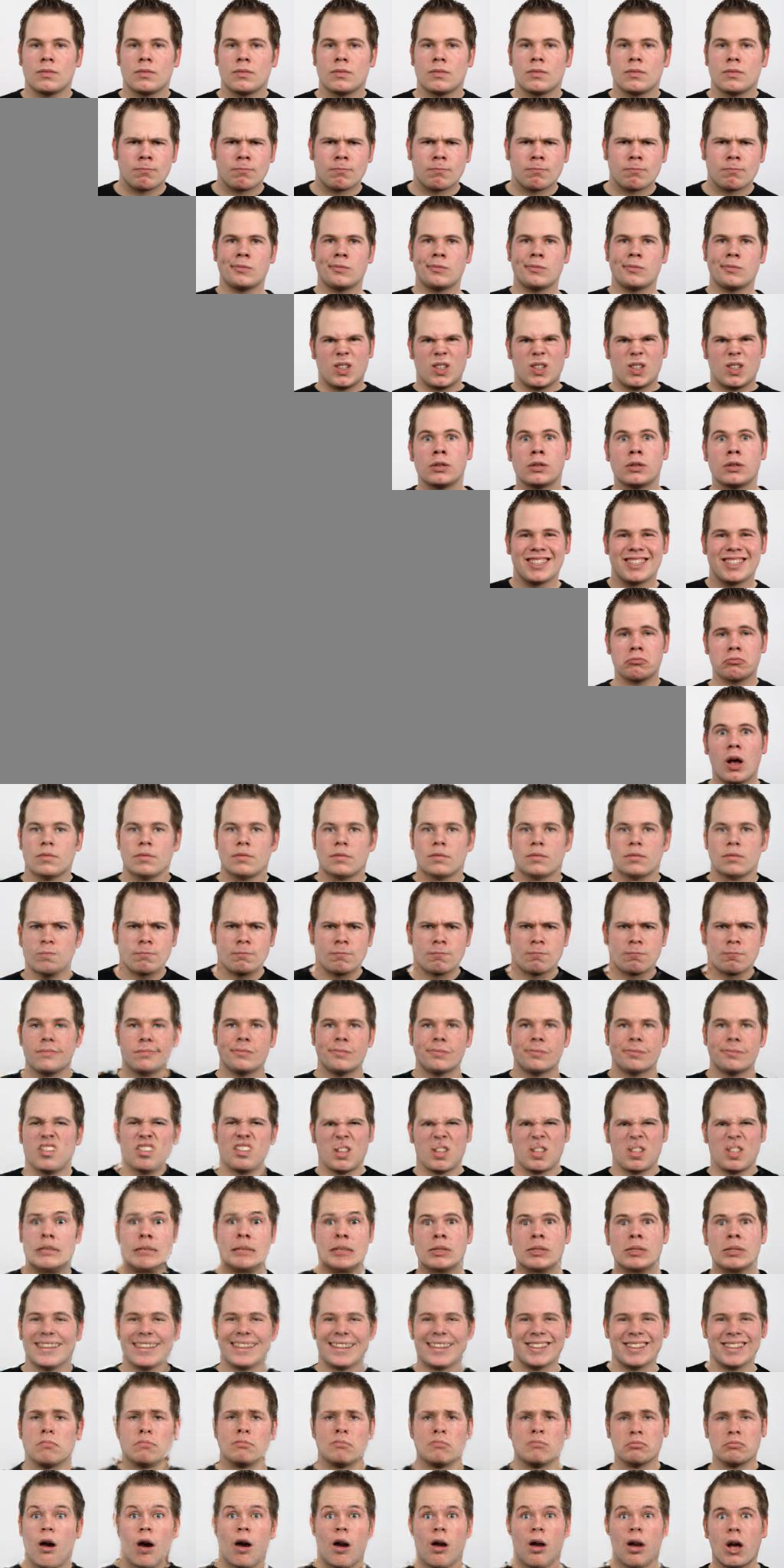}
  \caption{Random multi-domain image completion results of three testing samples in RaFD. The completed images (bottom) are generated from partial existing images in inputs (top). The number of inputted visible domains is in the range of $[1, N]$ where $N=8$ is the number of all domains. Rows: 8 image domains of ``neutral'', ``angry'', ``contemptuous'', ``disgusted'', ``fearful'', ``happy'', ``sad'', ``surprised''.}
\label{fig:face_n2n_2}
\end{figure*}

In addition to the qualitative results of image completion with multiple random missing modalities as shown in the Figs. 5(b), 6(b), 8 in the main text, we demonstrate more testing samples of the three datasets as shown in Figs.~\ref{fig:brats_n2n}-\ref{fig:face_n2n_2}. The left half or the top half of each figure shows the input domain(s), where the missing domains are filled up with zeros. The right half or the bottom half of each figure shows the image generation results for all the $N$ domains no matter whether it exists in the input domains or not. Firstly, no matter how many or which domains are visible in the input, the proposed model could generate images for all the $N$ domains including the missing ones in an one inference go. Especially for the missing domains, the domain-specific image characteristics are well captured although they do not appear in the input images. Comparing the generated images in the same domain, we could see that the domain-specific style and domain-shared content are all preserved well even when we limit the number of input visible domain to be only 1. In addition, when the number of visible domains increases, the content in each domain image is enhanced gradually and gets closer to the target image. This illustrates that our model is efficiently learning a better content code complementarily from multiple visible domains.

\section{Extended Ablative Study and Results for Missing-domain Segmentation}
Based on the results of missing-domain image completion, we show that our proposed method could go beyond image translation to solve the missing-domain segmentation problem. Specifically, our model learns efficient content representations of the subject, which could be efficiently leveraged for high-level recognition tasks. As shown in Fig. 3 in main text, a segmentation branch is added after the learned content code to generate segmentation prediction map. We adopt the dice loss as the segmentation loss in the training process. We run the segmentation experiments on both BraTS and ProstateX datasets, and use the dice score as the evaluation metric. In the following, we look into two specific settings in missing-domain segmentation.

\subsection{Missing-domain segmentation with inference on pre-trained segmentation model}
Suppose we have trained an oracle segmentation model on a complete dataset with all domain images. Then this pre-trained model would be used to predict segmentation results for new samples during the inference. For new subjects, some domains might be missing. Straightforward solutions to complete the missing domains include zero filling, average image computed from the existing domains, and the nearest neighbor (NN) searching among available training samples. We show the dice scores for these baseline methods in Table~\ref{tab:seg1}. Oracle results give the average testing dice score when all the domains are available in the inference. Each column shows the dice scores of segmentation predictions when the current domain is missing during inference. Moreover, based on image translation methods, we can generate fake images for missing domain imputation, and the results for different methods are shown in Table~\ref{tab:seg1}. We show that our proposed method achieves the best dice score compared with all aforementioned baselines and other GAN-based image translation methods. This also indicates that our method could generate better images by preserving a better content representation.
Furthermore, from the results in Table~\ref{tab:seg1}, we know that the T1Gd modality and the T2 modality are the most significant contrasts in the segmentation of BraTS and ProstateX data, missing of which will cause a severe performance decrease in dice score. Our method could alleviate such a loss to a large extent. 
Here, the dice score for BraTS is the average for the three segmentation categories: enhancing tumor (ET), tumor core (TC), and whole tumor (WT). Please see Table~\ref{tab:brats_seg1_full} for a full table with per-class dice scores.

\begin{table*}[t]
\centering
\caption{Missing-domain segmentation with inference on pre-trained segmentation model (average dice scores are reported)}
\label{tab:seg1}
\resizebox{0.8\textwidth}{!}{
\begin{tabular}{l|l|l|l|l|l|l|l}
\Xhline{2\arrayrulewidth}
\multicolumn{1}{c|}{Methods} & \multicolumn{4}{c|}{BraTS} & \multicolumn{3}{c}{ProstateX} \\
\cline{2-8}
\multicolumn{1}{c|}{} & \multicolumn{1}{c|}{T1} & \multicolumn{1}{c|}{T1Gd} & \multicolumn{1}{c|}{T2} & \multicolumn{1}{c|}{FLAIR} & \multicolumn{1}{c|}{T2} & \multicolumn{1}{c|}{ADC} & \multicolumn{1}{c}{HighB} \\
\Xhline{2\arrayrulewidth}
{Oracle} & \multicolumn{4}{c|}{0.822} & \multicolumn{3}{c}{0.908} \\ 
\Xhline{2\arrayrulewidth}
{Zero} & {0.651} & {0.473} & {0.707} & {0.454} & {0.528} & {0.243} & {0.775} \\ 
{Average} & {0.763} & {0.596} & {0.756} & {0.671} & {0.221} & {0.692} & {0.685} \\ 
{NN} & {0.769} & {0.540} & {0.724} & {0.606} & {0.759} & {0.850} & {0.854} \\
\cline{1-8}
{MUNIT} & {0.783} & {0.537} & {0.782} & {0.492} & {0.783} & {0.708} & {0.858} \\ 
{StarGAN} & {0.799} & {0.553} & {0.746} & {0.613} & {0.632} & {0.653} & {0.832} \\ 
{CollaGAN} & {0.753} & {0.564} & {0.798} & {0.674} & {0.472} & {0.760} & {0.842} \\
\Xhline{2\arrayrulewidth}
{ReMIC} & {\textbf{0.819}} & {\textbf{0.641}} & {\textbf{0.823}}& {\textbf{0.784}} & {\textbf{0.863}} & {\textbf{0.907}} & {\textbf{0.903}} \\
\Xhline{2\arrayrulewidth}
\end{tabular}}
\end{table*}

\begin{table*}[t]
\centering
\caption{Missing-domain segmentation with re-training segmentation model (average dice scores are reported)}
\label{tab:seg2}
\resizebox{0.8\textwidth}{!}{
\begin{tabular}{l|l|l|l|l|l|l|l}
\Xhline{2\arrayrulewidth}
\multicolumn{1}{c|}{Methods} & \multicolumn{4}{c|}{BraTS} & \multicolumn{3}{c}{ProstateX} \\
\cline{2-8}
\multicolumn{1}{c|}{} & \multicolumn{1}{c|}{T1} & \multicolumn{1}{c|}{T1Gd} & \multicolumn{1}{c|}{T2} & \multicolumn{1}{c|}{FLAIR} & \multicolumn{1}{c|}{T2} & \multicolumn{1}{c|}{ADC} & \multicolumn{1}{c}{HighB} \\
\Xhline{2\arrayrulewidth}
{Oracle} & \multicolumn{4}{c|}{0.822} & \multicolumn{3}{c}{0.908} \\ 
\Xhline{2\arrayrulewidth}
{Zero} & {0.811} & {0.656} & {0.823} & {0.775} & {0.868} & {0.899} & {0.897} \\
{Average} & {0.796} & {0.604} & {0.788} & {0.759} & {0.856} & {0.885} & {0.897} \\
\Xhline{2\arrayrulewidth}
{ReMIC} & {0.789} & {0.655} & {0.805} & {0.765} & {0.871} & {0.898} & {0.891} \\
{ReMIC+Seg} & {0.806} & {0.674} & {0.822} & {0.771} & {\textbf{0.872}} & {\textbf{0.909}} & {\textbf{0.905}} \\
{ReMIC+Joint} & {\textbf{0.828}} & {\textbf{0.693}} & {\textbf{0.828}} & {\textbf{0.791}} & {0.867} & {0.904} & {0.904} \\ 
\Xhline{2\arrayrulewidth}
\end{tabular}}
\end{table*}

\subsection{Missing-domain segmentation with re-training segmentation model}
Suppose we would like to train a segmentation model for a new data set, but most patients in this cohort just contain a random subset of all required domains. In this scenario, it is definitely not efficient to just use the most common domain overlapped by most patients. One simple solution is to complete all the missing images in training set by some imputation method, such as zero-filling image, average image, or generating images via image translation model. The results for these methods are shown in Table~\ref{tab:seg2}. More advanced, based on the content code learned in our model, we could develop a join model for multi-task learning of both generation and segmentation. 
By optimizing the generation loss and segmentation loss simultaneously, the unified model could learn how to generate missing images to promote segmentation performance. The results of jointly learned model as shown in Table~\ref{tab:seg2} achieve the best dice score in both BraTS and ProstateX datasets. ``ReMIC+Seg'' stands for using separate content encoders for generation and segmentation tasks, while ``ReMIC+Joint'' indicates sharing the weights of content encoder for the two tasks. We note that the baseline methods get better results after retraining the model on the missing data, since the model is trained to fit to the exact missing inputs format by optimizing the segmentation objective under the supervision of segmentation labels, which makes it more robust to missing inputs. However, our method can still get the best results through adaptive learning model. 

\begin{table*}[t]
\centering
\caption{Missing-domain segmentation with inference on pre-trained 2D and 3D segmentation model (per-class dice scores are reported)}
\label{tab:brats_seg1_full}
\resizebox{\textwidth}{!}{
\begin{tabular}{l|l|l|l|l|l}
\Xhline{2\arrayrulewidth}
\multicolumn{2}{c|}{Methods}    & \multicolumn{1}{c|}{T1} & \multicolumn{1}{c|}{T1Gd} & \multicolumn{1}{c|}{T2} & \multicolumn{1}{c}{FLAIR}
\\ \cline{3-6}
\multicolumn{2}{c|}{}  & \multicolumn{1}{c|}{WT / TC / ET} & \multicolumn{1}{c|}{WT / TC / ET} & \multicolumn{1}{c|}{WT / TC / ET} & \multicolumn{1}{c}{WT / TC / ET}
\\ 
\Xhline{2\arrayrulewidth}
{2D}
& {Oracle} & \multicolumn{4}{c}{0.910 / 0.849 / 0.708} \\ 
\cline{2-6} 
& {Zero} & {0.771 / 0.609 / 0.572} & {0.872 / 0.539 / 0.008} & {0.755 /  0.690 / 0.677} & {0.458 / 0.468 / 0.435} \\ 
& {Average} & {0.870 / 0.744 / 0.674} & {0.882 / 0.603 / 0.303} & {0.849 / 0.732 / 0.686} & {0.655 / 0.710 / 0.648} \\ 
& {NN} & {0.883 / 0.765 / 0.660} & {0.871 / 0.564 / 0.186} & {0.811 / 0.720 / 0.642} & {0.534 / 0.669 / 0.614} \\
\cline{2-6}
& {MUNIT} & {0.886 / 0.785 / 0.679} & {0.872 / 0.552 / 0.187} & {0.882 / 0.781 / 0.682} & {0.408 / 0.541 / 0.527} \\ 
& {StarGAN} & {0.897 / 0.795 / 0.704} & {0.886 / 0.588 / 0.184} & {0.851 / 0.725 / 0.661} & {0.570 / 0.664 / 0.604} \\ 
& {CollaGAN} & {0.860 / 0.747 / 0.651} & {0.864 / 0.576 / 0.252} & {0.882 / 0.811 / 0.700} & {0.663 / 0.697 / 0.663} \\
\cline{2-6}
& {ReMIC} & {\textbf{0.909} / \textbf{0.834} / \textbf{0.714}} & {\textbf{0.899} / \textbf{0.669} / \textbf{0.354}} & {\textbf{0.905} / \textbf{0.855} / \textbf{0.709}} & {\textbf{0.853} / \textbf{0.807} / \textbf{0.691}}
\\ 
\Xhline{2\arrayrulewidth}
{3D}
& {Oracle} & \multicolumn{4}{c}{0.909 / 0.867 / 0.733} \\
\cline{2-6} 
& {Zero} & {0.876 / 0.826 / 0.694} & {0.884 / 0.574 / 0.020} & {0.901 / 0.865 / 0.728} & {0.661 / 0.730 / 0.643} \\ 
& {Average} & {0.880 / 0.814 / 0.640} & {0.854 / \textbf{0.618} / \textbf{0.282}} & {0.838 / 0.801 / 0.695} & {0.713 / 0.732 / 0.675} \\ 
& {NN} & {0.890 / 0.829 / 0.703} & {0.859 / 0.538 / 0.081} & {0.790 / 0.799 / 0.704} & {0.472 / 0.686 / 0.607} \\ 
\cline{2-6}
& {ReMIC} & {\textbf{0.905} / \textbf{0.864} / \textbf{0.722}} & {\textbf{0.888} / 0.614 / 0.273} & {\textbf{0.902} / \textbf{0.871} / \textbf{0.734}} & {\textbf{0.855} / \textbf{0.850} /\textbf{ 0.724}} \\ 
\Xhline{2\arrayrulewidth}
\end{tabular}}
\end{table*}


\subsection{3D image segmentation with missing domains}
Furthermore, we validate that our method could not only work for 2D image segmentation but also 3D image segmentation. When a 3D volumetric image is missing in some domain, we deploy our method to generate 2D images per slice and stack them to build the whole 3D volumetric image in the corresponding missing domain. As shown in Table~\ref{tab:brats_seg1_full}, we evaluate the per-class dice score for missing-domain imputation with the oracle model trained from complete-domain 3D image segmentation. The results show our method could give a better performance in most domains. During experiments, we find that the smoothness among different slices in the 3D image generation might be an issue that needs to be further improved. 
Besides, we also show that the per-class dice scores for BraTS segmentation results in Table~\ref{tab:brats_seg1_full}. Compared with WT and TC classes, ET class is definitely more challenging in the brain tumor segmentation, since enhancing tumor usually just covers a very small region among the whole tumor. Particularly in the ET class segmentation, we can see our method outperforms the other methods to a large extent.

\subsection{Analysis of missing-domain segmentation results}
\begin{figure*}[h]
\centering
  \includegraphics[width=0.7\linewidth]{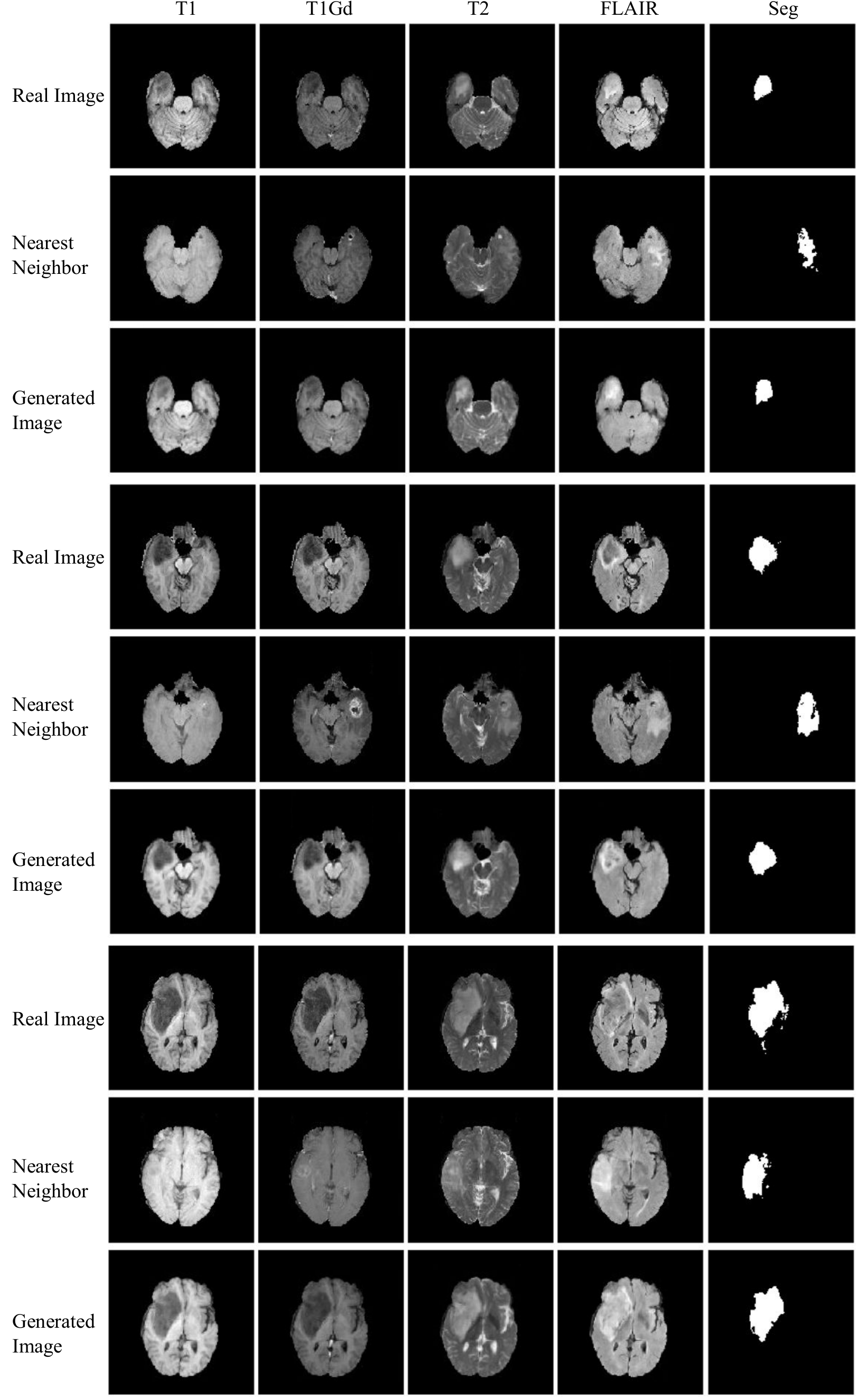}
  \caption{Missing-domain segmentation results of three testing samples in BraTS. Every three rows show results for one testing sample. For each testing sample, we show: 1) real images with ground truth segmentation label, 2) nearest neighbor searched from training data with its segmentation label, 3) generated images using our method and segmentation prediction when T1 image is missing and completed with the generated image.}
\label{fig:nn_seg_brats}
\end{figure*}

\begin{figure*}[h]
\centering
  \includegraphics[width=0.6\linewidth]{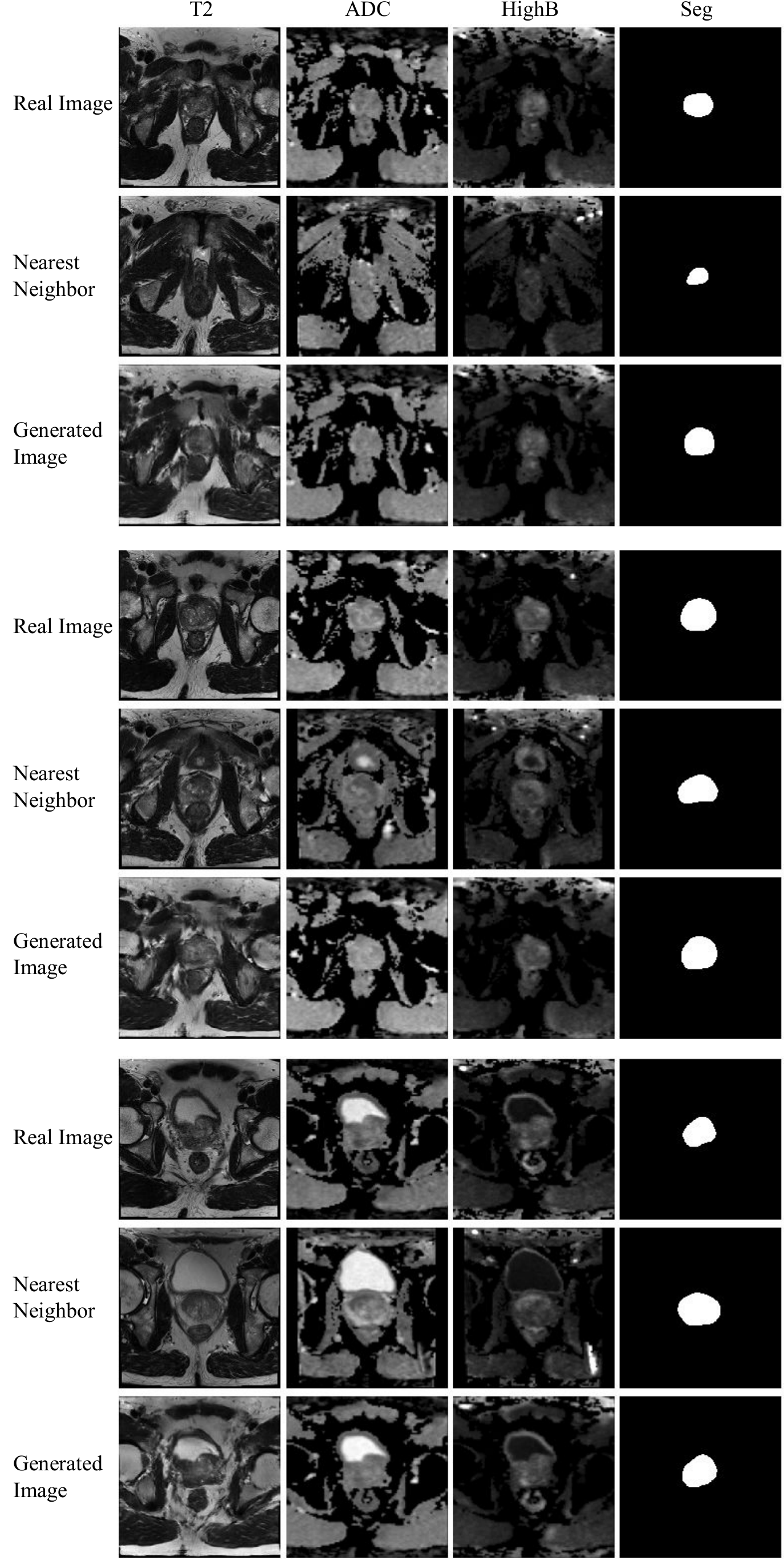}
  \caption{Missing-domain segmentation results of three testing samples in ProstateX. Every three rows show results for one testing sample. For each testing sample, we show: 1) real images with ground truth segmentation label, 2) nearest neighbor searched from training data with its segmentation label, 3) generated images using our method and segmentation prediction when T2 image is missing and completed with the generated image.}
\label{fig:nn_seg_pt}
\end{figure*}

To better understand why our method is a better solution in missing-domain imputation for multi-domain recognition tasks like the multi-modal image segmentation, we demonstrate three randomly selected testing samples in BraTS and ProstateX dataset as shown in Figs.~\ref{fig:nn_seg_brats}-\ref{fig:nn_seg_pt} respectively. Rows 1$\sim$3 shows the results for the first sample, and the other two samples are shown in the same format. In each sample, the first row shows real images in four domains and its ground truth segmentation labels. If some domain is randomly missing for the target sample, a straightforward solution is to search through all the available training data and find the nearest neighbor (NN) sample to complete the missing image. We search the nearest neighbor according to the Euclidean distance in 2D image space, and display the NN sample with all modalities, which actually looks very similar to the target sample visually. However, we note that the tumor region is seriously different between the target sample and its NN sample, which shows that the NN images is not a good missing imputation in terms of the image semantics. To cope with this issue, our proposed method is able to generate images for missing domains with not only pixel-level similarity but also similar predicted tumor regions, which are the most significant semantics in the tumor segmentation task. As shown in Figs.~\ref{fig:nn_seg_brats}-\ref{fig:nn_seg_pt}, the generated images in multiple domains closely resemble the target images. The segmentation map shows the prediction results when the generated T1 (T2) image is used as the imputation in inputs for BraTS (ProstateX) segmentation, which predicts a segmentation mask very close to the ground truth label. These results illustrate the superiority of our method in efficiently learning semantic content codes in the feature level.


\end{document}